%% file: main.tex
\def\year{2021}\relax
\documentclass[letterpaper]{article} % DO NOT CHANGE THIS
\usepackage{aaai21}  % DO NOT CHANGE THIS
\usepackage{times}  % DO NOT CHANGE THIS
\usepackage{helvet} % DO NOT CHANGE THIS
\usepackage{courier}  % DO NOT CHANGE THIS
\usepackage[hyphens]{url}  % DO NOT CHANGE THIS
\usepackage{graphicx} % DO NOT CHANGE THIS
\usepackage{natbib}  % DO NOT CHANGE THIS AND DO NOT ADD ANY OPTIONS TO IT
\usepackage{caption} % DO NOT CHANGE THIS AND DO NOT ADD ANY OPTIONS TO IT
\usepackage{amsmath}
\usepackage{amssymb}
\usepackage{booktabs}
\usepackage{algorithm}
\usepackage{algorithmic}

\input{notation.tex}

\title{Clusterability in Neural Networks}
\author {
    Daniel Filan\textsuperscript{\rm 1, *},
    Stephen Casper\textsuperscript{\rm 2, *},
    Shlomi Hod\textsuperscript{\rm 3, *}, \\
    Cody Wild\textsuperscript{\rm 1},
    Andrew Critch\textsuperscript{\rm 1},
    Stuart Russell\textsuperscript{\rm 1}\\
}
\affiliations {
    \textsuperscript{\rm 1} UC Berkeley,
    \textsuperscript{\rm 2} Harvard,
    \textsuperscript{\rm 3} Boston University \\
    \{daniel\_filan, codywild, critch, russell\}@berkeley.edu, scasper@college.harvard.edu, shlomi@bu.edu
}

\begin{document}

\maketitle

\begin{abstract}
The learned weights of a neural network have often been considered devoid of scrutable internal structure.
In this paper, however, we look for structure in the form of clusterability: how well a network can be divided into groups of neurons with strong internal connectivity but weak external connectivity.
We find that a trained neural network is typically more clusterable than randomly initialized networks, and often clusterable relative to random networks with the same distribution of weights.
We also exhibit novel methods to promote clusterability in neural network training, and find that in multi-layer perceptrons they lead to more clusterable networks with little reduction in accuracy.
Understanding and controlling
the clusterability of neural networks will hopefully render their inner workings more interpretable to engineers
by facilitating partitioning into meaningful clusters.
\end{abstract}

\noindent Modularity is a common property of biological and engineered systems \cite{clune2013evolutionary, baldwin2000design, booch2007object}.
Reasons for modularity include adaptability
and
the ability to handle different situations with common sub-problems.
It is also desirable from a perspective of transparency:
modular systems allow those analyzing the system to inspect the function of individual modules and combine their understanding of each into an understanding of the entire system.

In this work, we study a graph-theoretic analog to modularity:
the extent to which a network can be partitioned into sets of neurons where each set is strongly internally connected,
but only weakly connected to other sets.
This definition refers only to the learned weights of the network, 
not to the data distribution,
nor to the distributions of outputs or activations of the model.
More specifically, we use a spectral clustering algorithm \cite{shi2000normalized} to decompose trained networks into clusters,
and measure the goodness of this decomposition.
Since any degree of non-uniformity of weights can induce clusterability,
we also measure the relative clusterability of a network compared to networks with the same set of weights in each layer but shuffled randomly,
in order to
determine whether any clusterability is simply due to each layer's distribution of weights.

We conduct an empirical investigation into the clusterability of multi-layer perceptrons (MLPs) and convolutional neural networks (CNNs) trained on MNIST, Fashion-MNIST, and CIFAR-10 \cite{mnist, fashion_mnist, cifar}, using
weight pruning and other regularization methods. We also test if clusterability can be induced by training on datasets that benefit from some degree of parallelism to classify. In addition, we test the clusterability of neural networks trained by other researchers in the VGG, ResNet, and Inception families \cite{vgg, resnet, inception-v3} for ImageNet classification \cite{imagenet_cvpr09}.
Finally, we investigate two ways of training neural networks specifically to promote clusterability: regularizing for clusterability
and clusterable initialization.

\begin{figure}
    \centering
    \includegraphics[width=0.7\columnwidth]{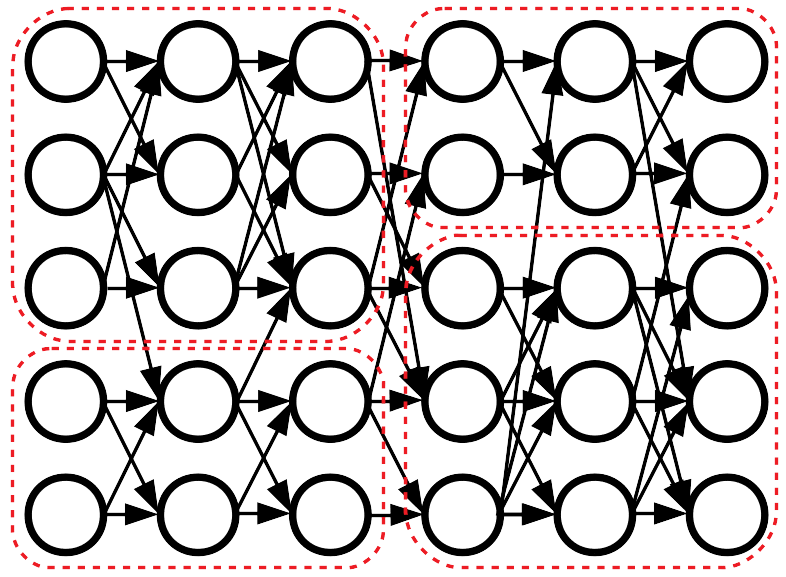}
    \caption{A pruned neural network, split into clusters.}
    \label{fig:clustered_net}
\end{figure}

Our main contributions are:
\begin{itemize}
    \item Presenting a definition of absolute and relative clusterability of a neural network (section~\ref{sec:defs_and_background}).
    \item Showing that trained neural networks are often more absolutely clusterable than randomly initialized networks (sections~\ref{sec:mlp_clusterability} and \ref{sec:cnn_clusterability_results}).
    \item Showing that neural networks trained with dropout and/or weight pruning are typically relatively clusterable, and often more clusterable than all 50 shuffled networks to which we compare them (sections~\ref{sec:mlp_clusterability} and \ref{sec:cnn_clusterability_results}).
    \item Showing that large neural networks trained for state-of-the-art ImageNet classification are reliably more clusterable than 99 out of 100 shuffled networks, and that a mid-sized VGG trained with dropout and weight pruning on CIFAR-10 classification reliably produces networks more clusterable than all 50 of their shuffles (section~\ref{sec:cnn_clusterability_results}).
    \item Demonstrating novel methods of promoting clusterability in MLPs in a way that is compatible with standard neural network architectures and training procedures, with little loss in accuracy (section~\ref{sec:encouraging-modularity}).
\end{itemize}

Code is available at \url{https://github.com/dfilan/clusterability_in_neural_networks}.

\section{Clustering Neural Networks}
\label{sec:defs_and_background}

\subsection{Definitions}
\label{subsec:defs}

We represent a neural network as a weighted, undirected graph $G$.
To do this for an MLP, we identify each neuron with any incoming or outgoing non-zero weights\footnote{When networks are pruned, often some neurons have all incident weights pruned away, leaving them with no functional role in the network. We ignore these neurons in order to run spectral clustering.}, 
including the pixel inputs and logit outputs,
with an integer between $1$ and $N$, where $N$ is the total number of neurons, and take the set of neurons to be the set $V$ of vertices in $G$.
Two neurons have an undirected edge between them if they are in adjacent layers, and the weight of the edge is equal to the absolute value of the weight of the connection between the two neurons. We represent the set of weights by the adjacency matrix $A$ defined by $A_{ij} = A_{ji} :=$ the edge weight between neurons $i$ and $j$. If there is no edge between $i$ and $j$, then $A_{ij} = A_{ji} := 0$. As such, $A$ encodes all the weight matrices of the neural network, but not the biases.

For a CNN, the `neurons' we use are channels of the hidden layers---we omit the input layer and fully-connected layers at the end of the network.
In this case, the weight of the edge between two channels is the $L_1$ norm of the two-dimensional slice of the convolutional filter by which the input channel maps to the output channel.\footnote{We also tried constructing the graph with the $L_2$ norm, and found essentially similar results.}
When batch normalization is used between two convolutional layers, we divide weights by the moving standard deviation
and multiply them by the scaling factor $\gamma$.

The degree of a neuron $i$ is defined by $d_i := \sum_j A_{ij}$. The degree matrix $D$ is a diagonal matrix where the diagonal elements are the degrees: $D_{ii} := d_i$. 
We define the volume of a set of neurons $X \subseteq V$ as $\vol(X) := \sum_{i \in X} d_i$, 
and the weight between two disjoint sets of neurons $X,Y \subseteq V$ as $W(X,Y) := \sum_{i \in X, j \in Y} A_{ij}$. If $X \subseteq V$ is a set of neurons, then we denote its complement as $\bar{X} := V \setminus X$.

A partition of the network is a collection of disjoint subsets $X_1, \dotsc, X_k \subseteq V$
whose union forms the whole vertex set.
Our `goodness measure' of a partition is the normalized cut metric \cite{shi2000normalized} defined as $\ncut(X_1, \dotsc, X_k) := \sum_{i = 1}^k W(X_i, \bar{X_i}) / \vol(X_i)$, which we call `n-cut' in text.
The n-cut will be low if neurons in the same partition element tend to share high-weight edges and those in different partition elements share low-weight edges or no edges at all,
as long as the sums of degrees of neurons in each partition element are roughly balanced.\footnote{For a probabilistic interpretation of n-cut that gives more intuitive meaning to the quantity, see appendix \ref{app:ncut_meaning}.}

Finally, the graph Laplacian 
is defined as $L := D - A$,\footnote{For the connection to the second derivative operator on $\mathbb{R}^n$, see \citet{laplacian_explanation} and \citet{von2007tutorial}.} 
and the normalized Laplacian as $\Lnorm := D^{-1}L$. $\Lnorm$ is a positive semi-definite matrix with $N$ real-valued non-negative eigenvalues \cite{von2007tutorial}.
The eigenvectors and eigenvalues of $\Lnorm$ are the generalized eigenvectors and eigenvalues of the generalized eigenvalue problem $L u = \lambda D u$.

\subsection{Spectral Clustering}
\label{subsec:spectral_clustering}

To measure the clusterability of a graph, we use a spectral clustering algorithm to compute a partition, which we call a clustering, and evaluate the n-cut.
The 
algorithm we use \cite{shi2000normalized} solves a relaxation of the NP-hard problem of finding a clustering that minimizes the n-cut \cite{von2007tutorial}.
It is detailed in algorithm~\ref{alg:spectral_clustering}, which is adapted from \citet{von2007tutorial}. We use the scikit-learn implementation \cite{scikit-learn} using
the ARPACK eigenvalue solver \cite{lehoucq1998arpack}.

\begin{algorithm}[tb]
\caption{Normalized Spectral Clustering}
\label{alg:spectral_clustering}
\begin{algorithmic}
\STATE {\bfseries Input:} Adjacency matrix $A$, number $k$ of clusters
\STATE Compute the normalized Laplacian $\Lnorm$
\STATE Compute the first $k$ eigenvectors $u_1, \dotsc, u_k \in \R^N$ of $\Lnorm$
\STATE Form the matrix $U \in \R^{k \times N}$ whose $j$\ts{th} row is $u_j^\top$
\STATE For $n \in \{1, \dotsc, N\}$, let $y_n \in \R^k$ be the $n$\ts{th} column of $U$
\STATE Cluster the points $(y_n)_{n=1}^N$ with the $k$-means algorithm into clusters $C_1, \dotsc, C_k$
\STATE {\bfseries Return:} Clusters $X_1, \dotsc, X_k$ with $X_i = \{n \in \{1, \dotsc, N\} \mid y_n \in C_i \}$
\end{algorithmic}
\end{algorithm}

We define the n-cut of a network as the n-cut of the clustering this algorithm~\ref{alg:spectral_clustering} returns, run with $k=12$.\footnote{
Appendix~\ref{app:n_clusters} shows results of some clusterability experiments for $k\in \{2,4,7,10\}$. Results are similar for all values except $k=2$.} 
Since the n-cut is low when the network is
clusterable, we will describe a decrease in n-cut as an increase in \emph{absolute clusterability}
and vice versa.

To measure the \emph{relative clusterability} of an MLP, we sample 50 random networks by randomly shuffling the weight matrix of each layer of the trained network.
We convert these networks to graphs, cluster them, and find their n-cuts. 
For CNNs, we shuffle the edge weights between channels once the network has been turned into a graph, which is equivalent to
shuffling which two channels are connected by each
spatial
kernel slice.
We then compare the n-cut of the trained network to the sampled n-cuts, estimating the left one-sided $p$-value \cite{north2002note}
and the Z-score: the number of standard deviations the network's n-cut lies below or above the mean of the shuffle distribution.
This determines whether the trained network is more clusterable than one would predict based only on its sparsity and
weight distribution.

\section{Clusterability in MLPs}
\label{sec:mlp_clusterability}

In this section, we report the results of experiments designed to determine the degree of clusterability of MLPs. For each experiment we train an MLP with 4 hidden layers, each of width 256, using Adam \cite{kingma2014adam}. 
After the network has neared convergence, we train for additional epochs with weight pruning on a polynomial decay schedule \cite{zhu2017prune} up to 90\% sparsity.
Pruning is used since the pressure to minimize connections plausibly causes modularity in biological systems \cite{clune2013evolutionary}. For further details on the training method, 
see appendix~\ref{app:train_hypers}.

First, we show results for MLPs trained on the MNIST and Fashion-MNIST datasets. We investigate networks trained with either no regularization, dropout with $p=0.5$, $L_1$ regularization with weight $5 \times 10^{-5}$, or $L_2$ regularization with weight $5 \times 10^{-5}$. For each condition, we train for 5 runs, and check clusterability both right before pruning as well as at the end of training. In all conditions, networks train to $\sim$98\% test accuracy on MNIST, and 87-89\% test accuracy on Fashion-MNIST. At initialization, these networks have n-cuts of between 10.2 and 10.4, as plotted in figure~\ref{fig:mlp_init_ncuts}.

When training with $L_2$ regularization, we occasionally found that networks had near-zero n-cut. This seemed to be due to extremely unbalanced clusterings where small groups of neurons were effectively disconnected from the rest of the network. When this happened, we retrained until the n-cut was significantly above zero, since we felt that the near-zero n-cuts did not reflect the true clusterability of the bulk of the network.

\begin{figure}
    \centering
    \includegraphics[width=0.9\columnwidth]{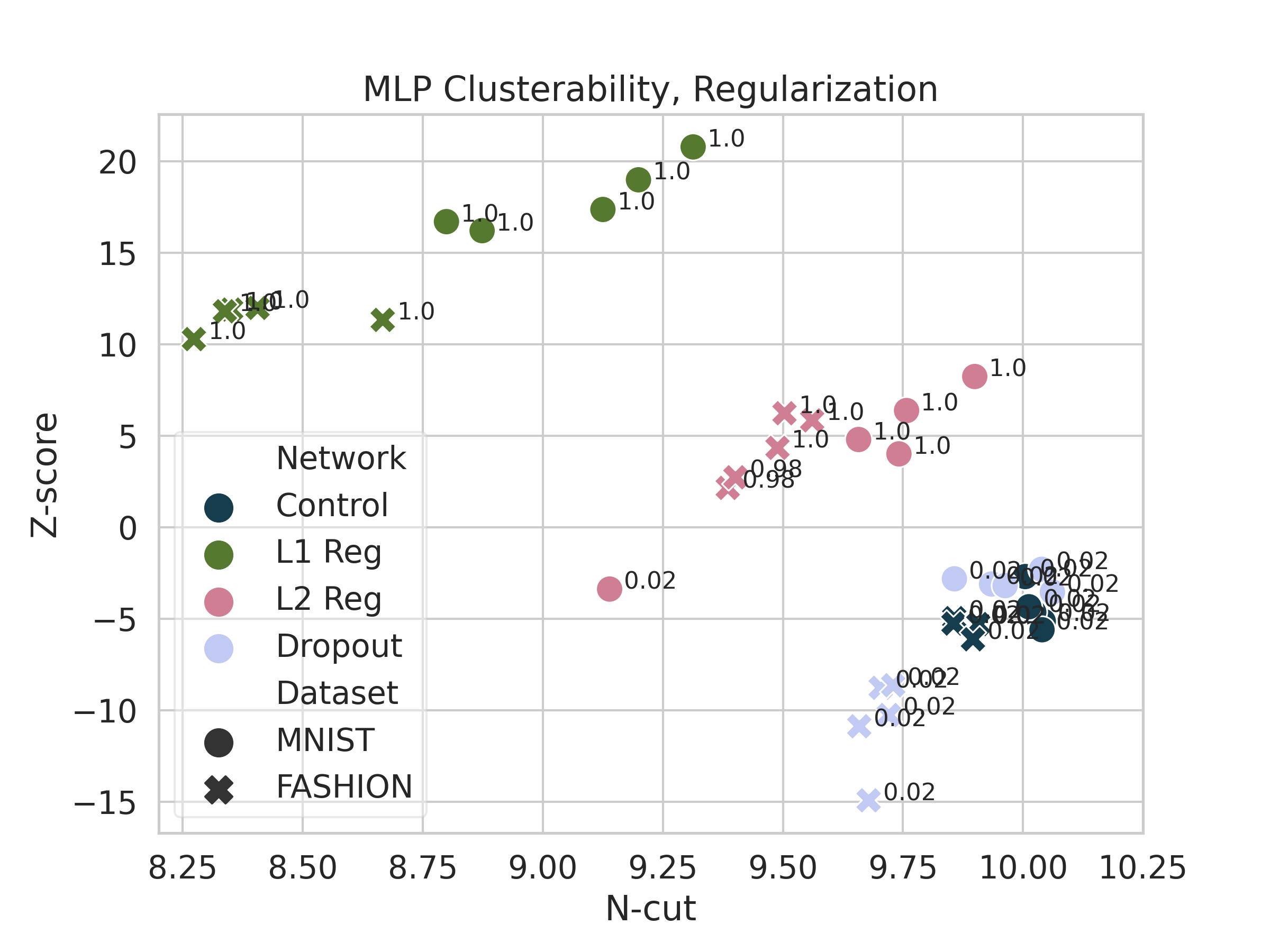}
    \caption{Clusterability of MLPs trained without pruning. Points are labeled with their one-sided $p$-value.}
    \label{fig:mlp_clusterability_unpruned}
\end{figure}

\begin{figure}
    \centering
    \includegraphics[width=0.9\columnwidth]{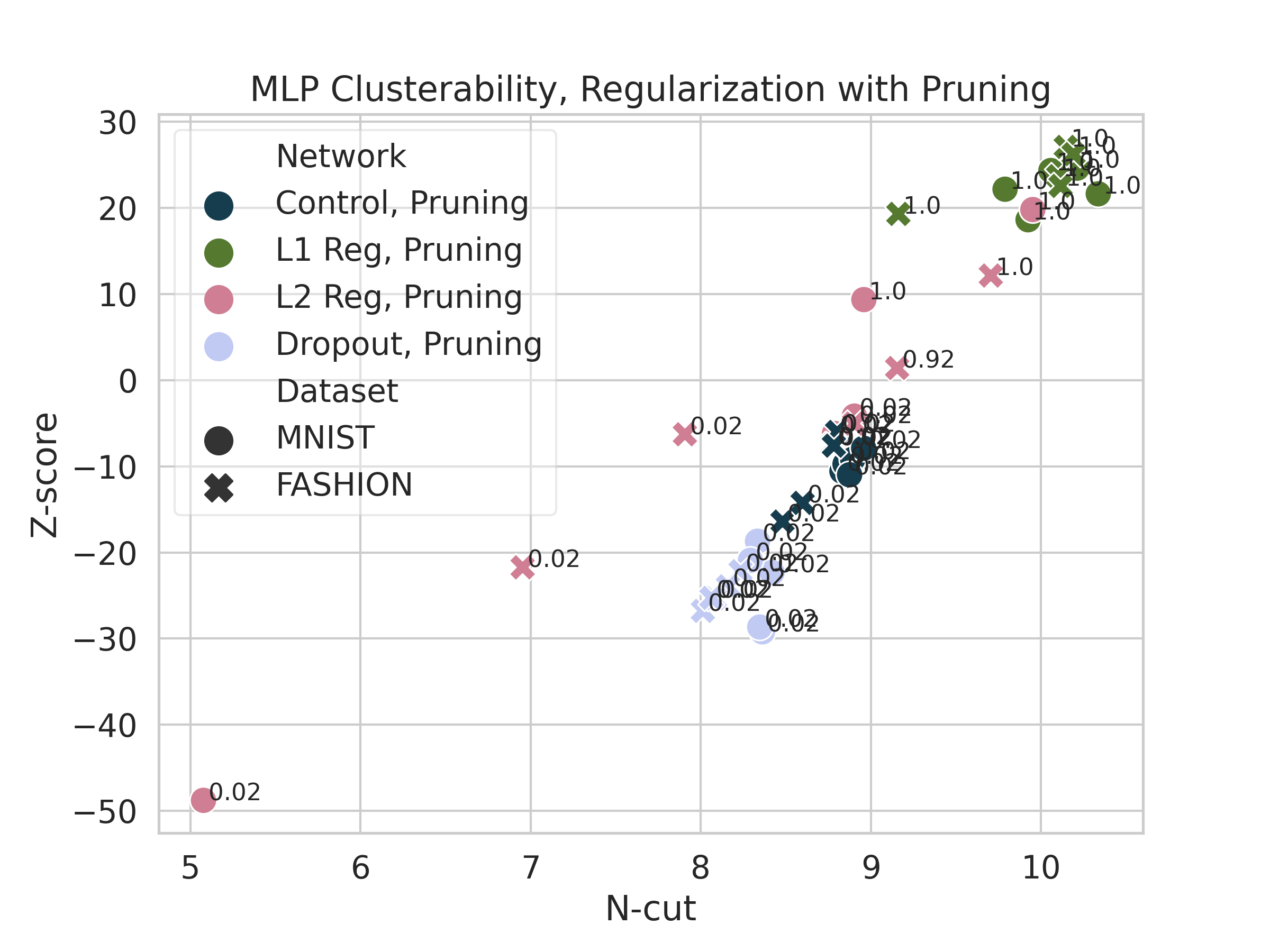}
    \caption{Clusterability of MLPs trained with pruning. Points are labeled with their one-sided $p$-value.}
    \label{fig:mlp_clusterability_pruned}
\end{figure}

Results are shown in figures~\ref{fig:mlp_clusterability_unpruned} and \ref{fig:mlp_clusterability_pruned}, and tables~\ref{tab:mlp_mnist_stats} and \ref{tab:mlp_fashion_stats}. We see that pruning promotes absolute clusterability, and dropout promotes absolute and relative clusterability. $L_1$ and $L_2$ regularization promote absolute clusterability pre-pruning, but not after pruning, at the expense of relative clusterability.
All networks appear to be more clusterable than at initialization, except those trained with $L_1$ regularization and pruning.

\subsection{`Halves' Datasets}
\label{subsec:mlp-modular-datasets}

We hypothesized that clusterability comes from different parts of the network independently computing different things. To test this, we trained on datasets that we hypothesized would lend themselves to
parallel processing.

\begin{figure}[ht]
    \centering
    \includegraphics[width=0.9\columnwidth]{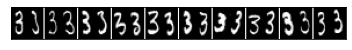}
    \includegraphics[width=0.9\columnwidth]{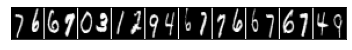}
    \includegraphics[width=0.9\columnwidth]{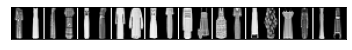}
    \includegraphics[width=0.9\columnwidth]{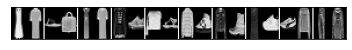}
    \caption{Samples from `halves' datasets, all of class 3. Each row has 10 images from the respective dataset. First row is MNIST-halves-same, second is MNIST-halves-diff, third is Fashion-halves-same, fourth is Fashion-halves-diff.}
    \label{fig:halves}
\end{figure}

The first type of dataset, called `MNIST-halves', features two MNIST images side-by-side, each shrunk in width so that the combined image is still $28 \times 28$ pixels.
In `MNIST-halves-same', both images are elements of the same class, and the class of the composite image is the class that the halves have.
By contrast, in `MNIST-halves-diff', the images come from random classes, and the class of the composite image is the sum of the classes of each half, modulo 10. 
For this task, it would be advantageous if the network could devote some neurons to processing one half, some to processing the other, and then combine the results.
Since the neurons processing one half would not need information from those processing the other, we speculate that networks trained on `halves-diff' will be more clusterable than those trained on those trained on `halves-same'.
We also make the same construction out of the Fashion-MNIST dataset, using the numerical class labels associated with the dataset.
Samples from these datasets are shown in figure~\ref{fig:halves}.
Networks train to around 99\% accuracy on MNIST-halves-same, 92\% on MNIST-halves-diff, 93-94\% on Fashion-halves-same, and 71-72\% on Fashion-halves-diff.

\begin{figure}
    \centering
    \includegraphics[width=0.9\columnwidth]{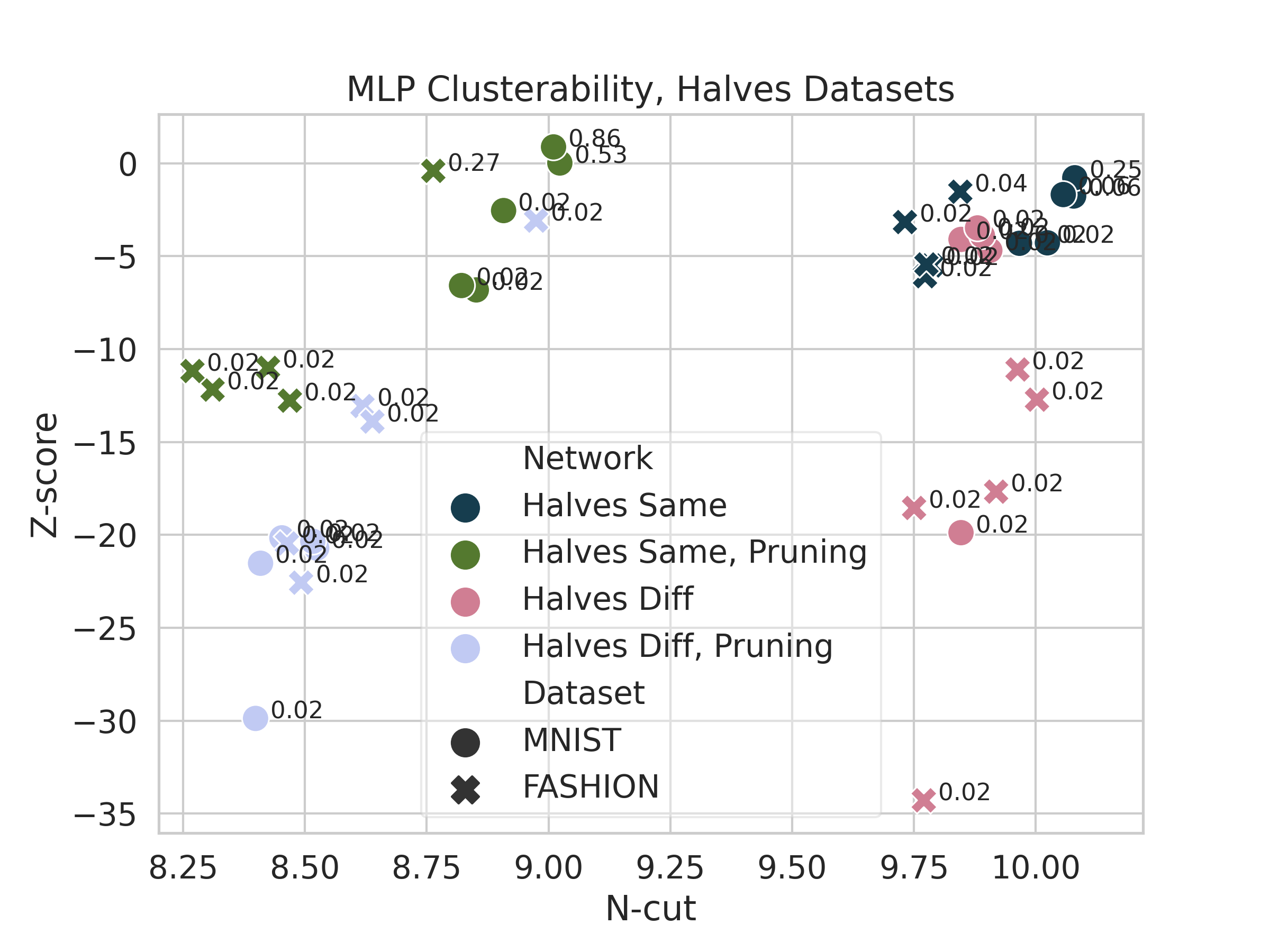}
    \caption{Clusterability of MLPs trained on `halves' datasets. Points are labeled with their one-sided $p$-value.}
    \label{fig:mlp_halves_clusterability}
\end{figure}

Results are shown in figure~\ref{fig:mlp_halves_clusterability} and table~\ref{tab:mlp_halves_stats}. Networks trained on halves-diff datasets are more relatively clusterable than those trained on halves-same datasets, but not more absolutely clusterable. All networks are more clusterable than at initialization.

We also find that training to regress multiple polynomials induces clusterability: 
for details, see appendix~\ref{app:poly_regression}. However, training on a dataset that is a simple mixture of two other datasets has ambiguous results, as detailed in appendix~\ref{app:mixture_datasets}. 
 
In addition, we ran experiments to explore the reasons for clusterability in pruned MLPs. In appendix~\ref{app:random_dataset_experiments} we outline experiments varying whether networks can or cannot memorize a dataset of random images to show that clusterability develops in tandem with learning, and in appendix~\ref{app:topology_preserving_shuffles} we determine that clusterability is not just due to the network topology, but also the particular arrangement of weights.

\section{Clusterability in CNNs}
\label{sec:cnn_clusterability_results}

To test if the results found on MNIST and Fashion-MNIST in section~\ref{sec:mlp_clusterability} are specific to the MLP architecture, we repeat them using a small CNN. Our network has 3 convolutional hidden layers of 64 channels each, followed by a fully-connected hidden layer with 128 neurons (further details on the architecture in appendix section~\ref{app:train_hypers}). $L_1$ and $L_2$ regularization strength is the same as for MLPs, and dropout rate is $0.25$ for convolutional layers and $0.5$ for fully-connected layers. In all regularization schemes, networks trained to around 99\% accuracy on MNIST and 90-92\% accuracy on Fashion-MNIST. At initialization, their n-cuts are between 10.90 and 10.95, as shown in figure~\ref{fig:small_cnn_init_ncuts}.

\begin{figure}
    \centering
    \includegraphics[width=0.9\columnwidth]{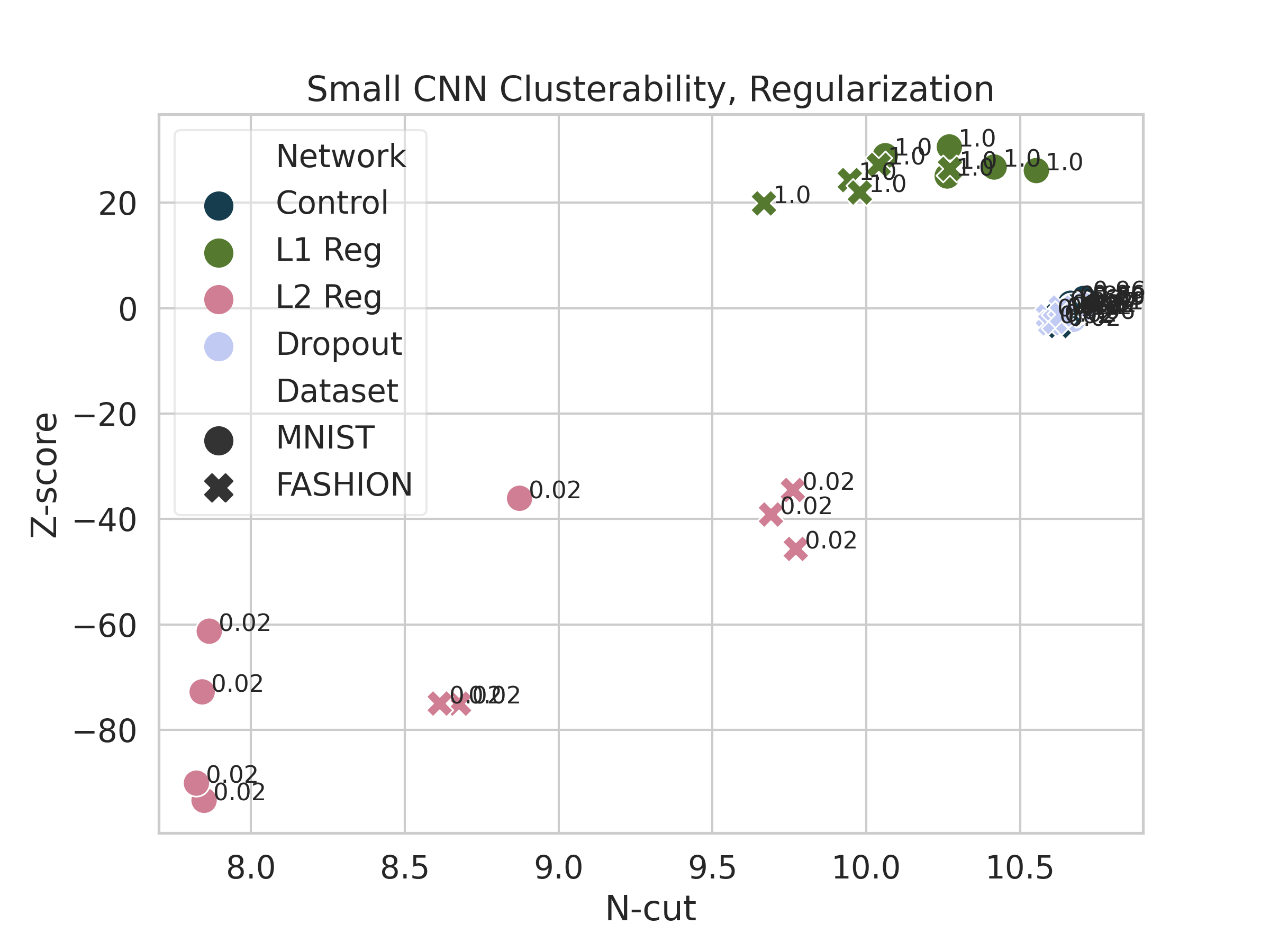}
    \caption{Clusterability of small CNNs trained without pruning. Points are labeled with their one-sided $p$-value. For detail on networks trained with no regularization or dropout, see figure~\ref{fig:small_cnn_unpruned_focus}.}
    \label{fig:small_cnn_unpruned}
\end{figure}

\begin{figure}
    \centering
    \includegraphics[width=0.9\columnwidth]{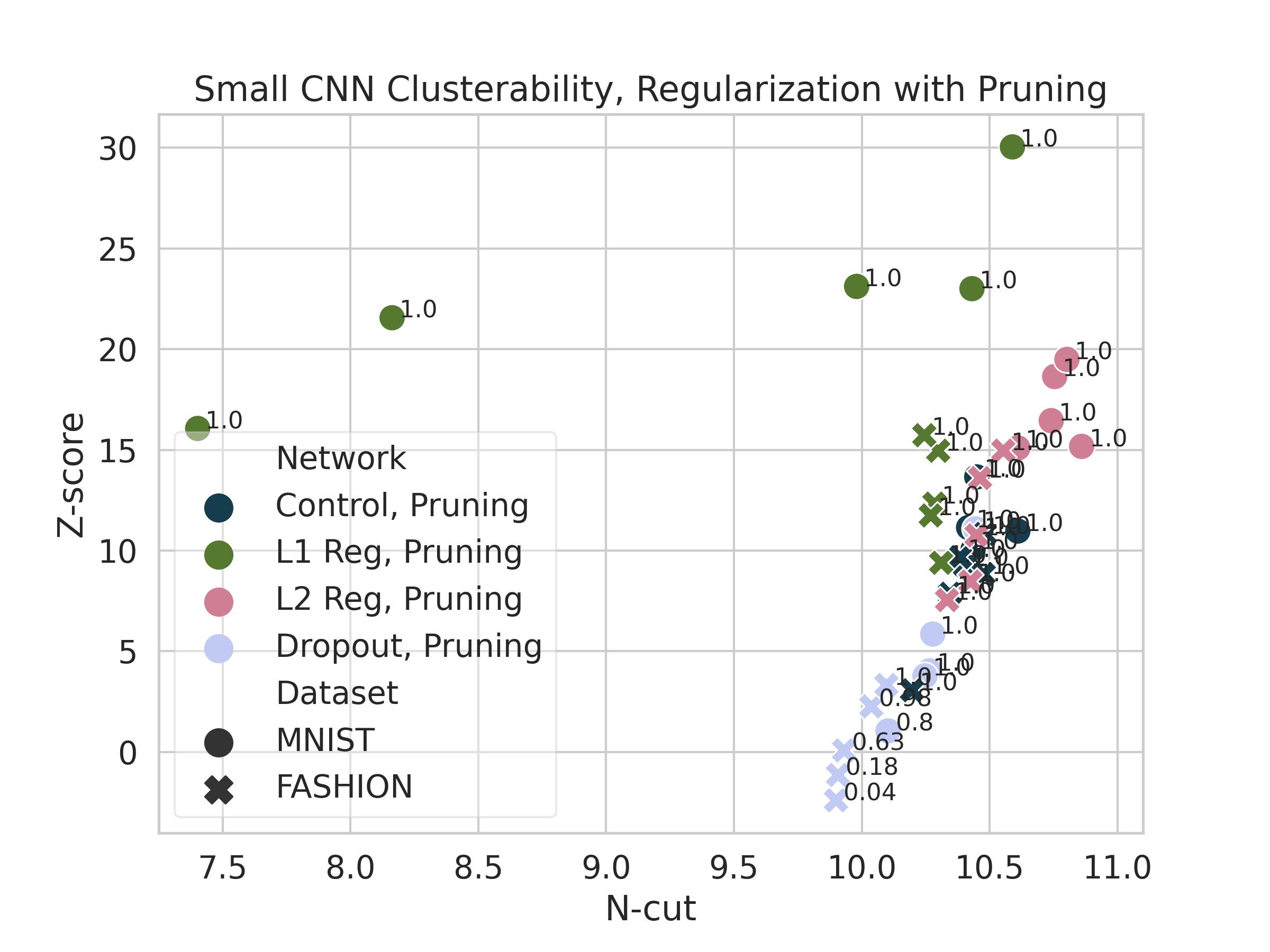}
    \caption{Clusterability of small CNNs trained with pruning. Points are labeled with their one-sided $p$-value.}
    \label{fig:small_cnn_pruned}
\end{figure}

Results are shown in figures~\ref{fig:small_cnn_unpruned} and \ref{fig:small_cnn_pruned}, and tables~\ref{tab:small_cnn_mnist_stats} and \ref{tab:small_cnn_fashion_stats}. We see that pruning fails to promote relative clusterability at all, that $L_1$ regularization promotes absolute but not relative clusterability, and that before pruning, $L_2$ regularization promotes absolute and relative clusterability. Networks are reliably more clusterable than at initialization, except those trained with $L_2$ regularization after pruning.

\begin{figure}
    \centering
    \includegraphics[width=0.9\columnwidth]{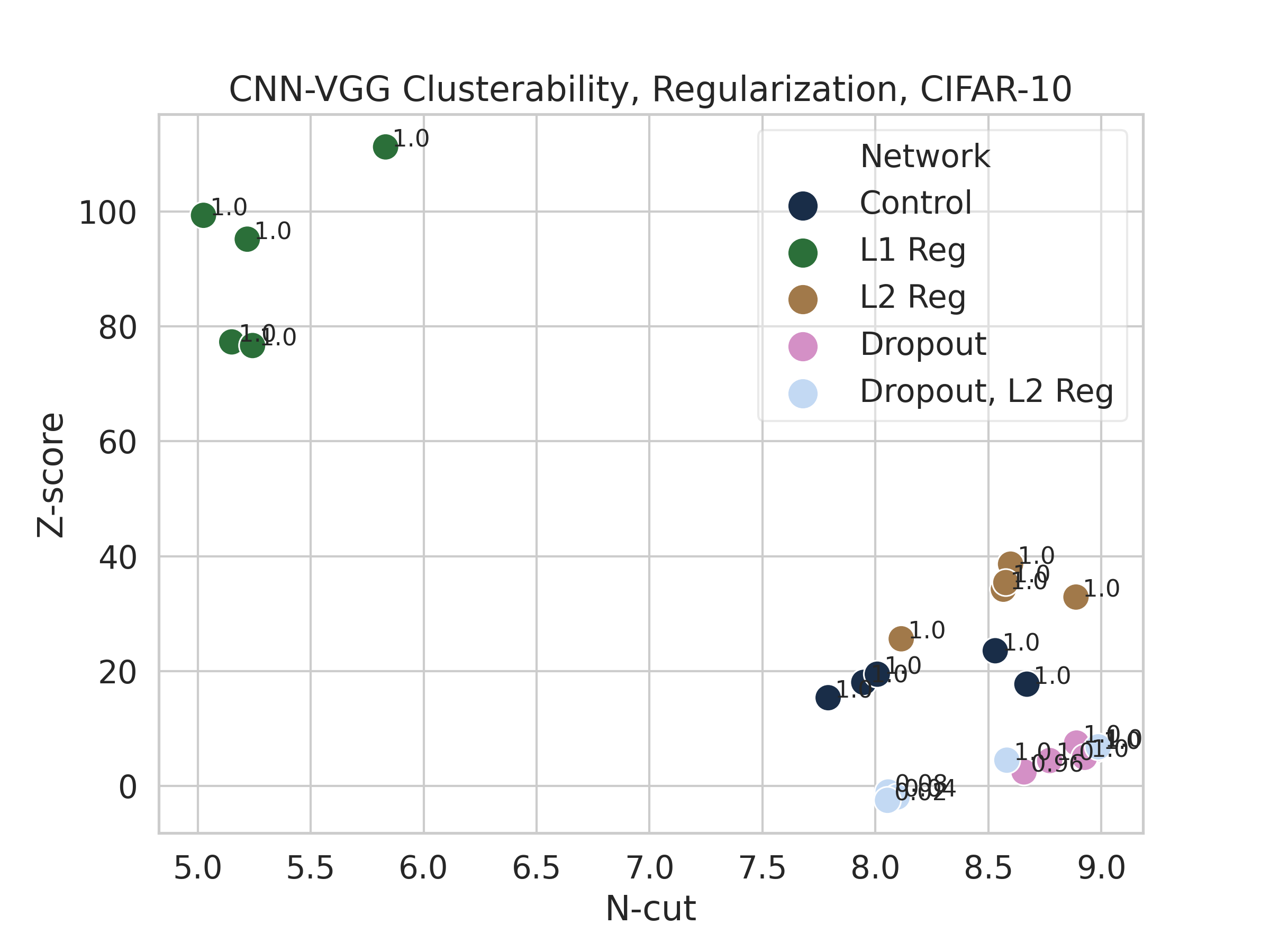}
    \caption{Clusterability of VGGs trained without pruning on CIFAR-10. Points are labeled with their one-sided $p$-value.}
    \label{fig:vgg_unpruned}
\end{figure}

\begin{figure}
    \centering
    \includegraphics[width=0.9\columnwidth]{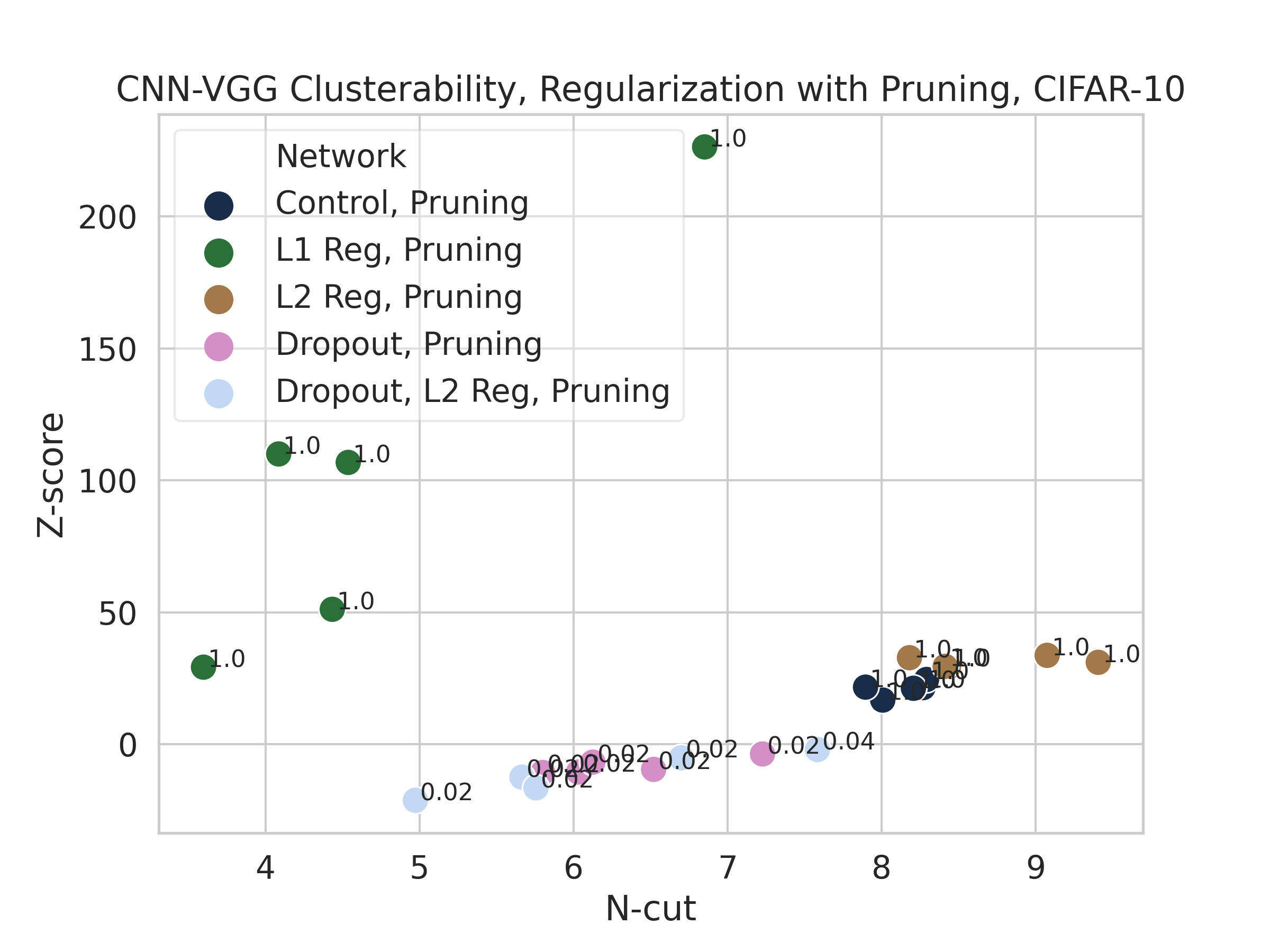}
    \caption{Clusterability of VGGs trained with pruning on CIFAR-10. Points are labeled with their one-sided $p$-value.}
    \label{fig:vgg_pruned}
\end{figure}

To see if these results generalize to larger networks on more complex datasets, we run the same experiments using a version of VGG-16 described by Liu and Deng \shortcite{DBLP:conf/acpr/LiuD15} trained on CIFAR-10. Data augmentation is used as described in appendix section~\ref{app:train_hypers}. We use a per-layer dropout rate as specified in Liu and Deng \shortcite{DBLP:conf/acpr/LiuD15}. Without pruning, we achieve 86-90\% test accuracy, and with pruning, we achieve 88-91\% test accuracy. At initialization, these networks have n-cuts between 8.45 and 8.63, as plotted in figure~\ref{fig:vgg_init_ncuts}. Results are shown in figures~\ref{fig:vgg_unpruned} and \ref{fig:vgg_pruned}, and table~\ref{tab:cifar_vgg_stats}.

We see that these CNNs are typically not relatively clusterable, except when trained with
dropout alone and pruning, or Liu and Deng's \shortcite{DBLP:conf/acpr/LiuD15} combination of dropout and $L_2$ regularization as well as pruning.
Networks trained without pruning are not more clusterable than at initialization, but networks trained with pruning are except when only $L_2$ regularization is used.

\subsection{Clusterability of ImageNet Models}
\label{subsec:imagenet-clusterability}

\begin{figure}
    \centering
    \includegraphics[width=0.9\columnwidth]{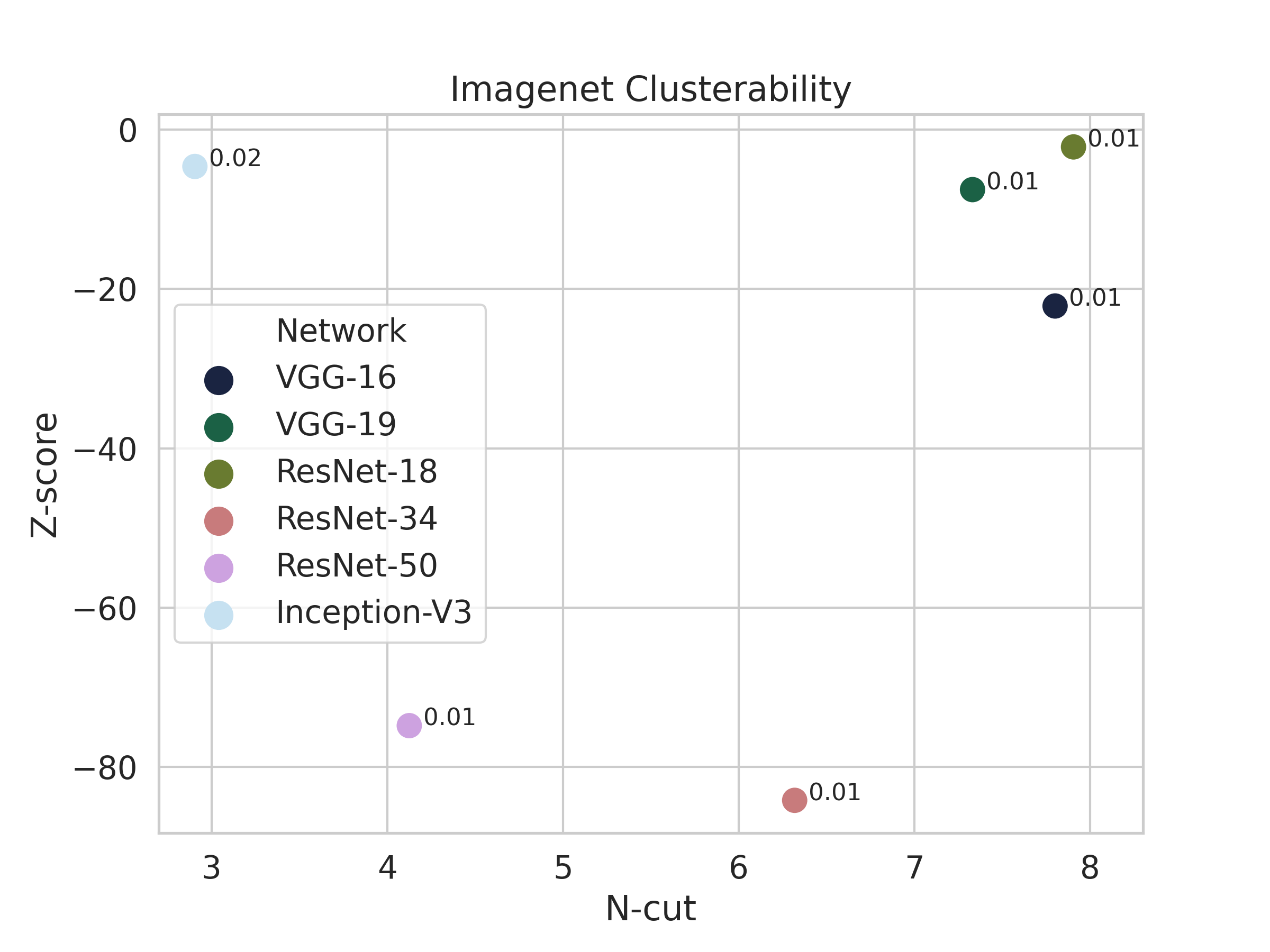}
    \caption{Clusterability of models trained on ImageNet. Points are labeled with their one-sided $p$-value.}
    \label{fig:imagenet_scatter}
\end{figure}

We also explore the clusterability of networks trained on ImageNet \cite{imagenet_cvpr09}. Specifically, we looked at VGG-16 and 19 \cite{vgg}, ResNet-18, 34, and 50 \cite{resnet}, and Inception-V3 \cite{inception-v3}. Weights were obtained from the Python \texttt{image-classifiers} package, version 1.0.0. Clustering was less stable for these networks, likely because of their large size, so
we used 100 random starts for $k$-means clustering instead of the default 10,
and models were compared to a distribution of 100 shuffles. Results are shown in figure~\ref{fig:imagenet_scatter} and table~\ref{tab:imagenet_table}. As we see, Inception-V3 is quite clusterable in an absolute sense, likely due to its modular architecture, and all networks are relatively clusterable.

\subsection{`Stack' Datasets}
\label{subsec:cnn-modular-datasets}

\begin{figure}[ht]
    \centering
    \includegraphics[width=0.9\columnwidth]{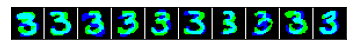}
    \includegraphics[width=0.9\columnwidth]{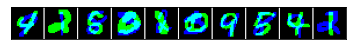}
    \includegraphics[width=0.9\columnwidth]{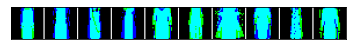}
    \includegraphics[width=0.9\columnwidth]{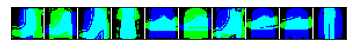}
    \caption{Samples from `stack' datasets, all of class 3. Each row has 10 images from the respective dataset. One channel is colored blue, the other is colored green. First row is MNIST-stack-same, second is MNIST-stack-diff, third is Fashion-stack-same, fourth is Fashion-stack-diff.}
    \label{fig:stack}
\end{figure}

As in subsection~\ref{subsec:mlp-modular-datasets}, we explored the effects of datasets we designed to induce modularity. Instead of the halves datasets, we used `stack' datasets where different images are in different input channels, rather than shrunk and put next to each other in one channel, because convolutional layers followed by max-pooling promote spatial location invariance within channels. The datasets are otherwise the same: in `stack-same', the channels are images of the same class, while in `stack-diff', the channels are of different classes and the label is the classes' sum modulo 10. Images are shown in figure~\ref{fig:stack}. We train the same CNN as was trained on MNIST and Fashion-MNIST.

\begin{figure}
    \centering
    \includegraphics[width=0.9\columnwidth]{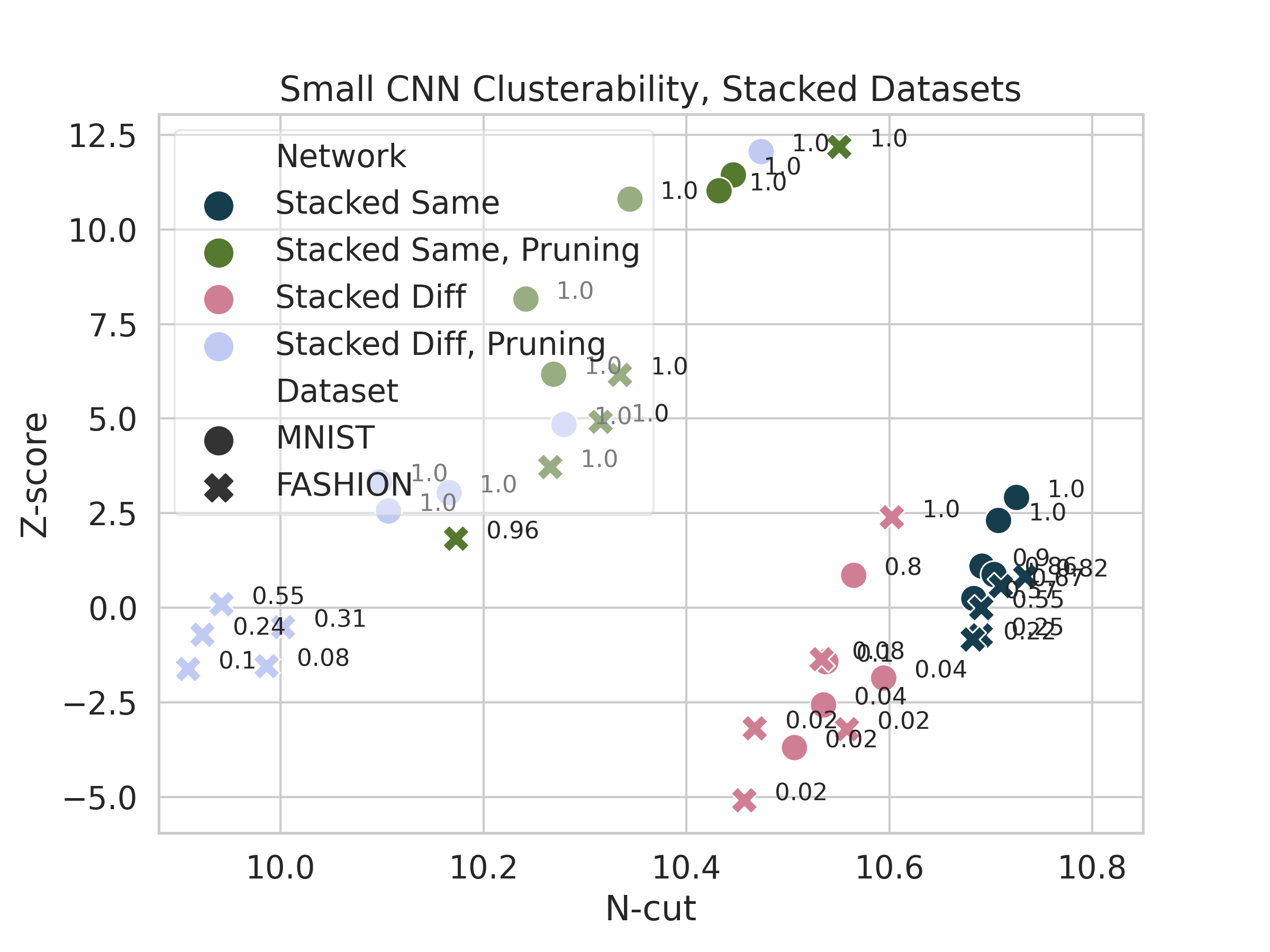}
    \caption{Clusterability of models trained on `stack' datasets. Points are labeled with their one-sided $p$-value.}
    \label{fig:cnn_stack}
\end{figure}

Results are shown in figure~\ref{fig:cnn_stack} and table~\ref{tab:small_cnn_stack_stats}. We see that pruning promotes absolute but not relative clusterability, while networks trained on stack-diff datasets are somewhat more clusterable, both in absolute and relative terms, than those trained on stack-same datasets. They are always more clusterable than at initialization.

\section{Promoting Clusterability}
\label{sec:encouraging-modularity}

We explore two methods to promote clusterability while hewing closely to standard training procedure: regularization and initialization.

\subsection{Regularization}
\label{subsec:clusterability_regularization}

The most straightforward approach to promote clusterability is regularization.
The Cheeger inequalities \cite{dodziuk1984difference, alon1985lambda1} give a bound on the Cheeger constant of a graph (closely related to n-cut for $k=2$) in terms of the second eigenvalue of
the normalized Laplacian matrix 
$\Lnorm$: namely, the lower the eigenvalue, the more easy it is to divide the network into two. Lee et al.\ \shortcite{lee2014multiway} have shown that this bound extends to the analog of the Cheeger constant for $k > 2$. As such, regularizing the $k$\ts{th} eigenvalue should produce a network with a low n-cut value for $k$ clusters.

This is possible because if a symmetric $n \times n$ matrix $S$ has distinct eigenvalues $\lambda_1, \dotsc, \lambda_n$ associated with orthonormal eigenvectors $u_1, \dotsc, u_n$, the derivative of $\lambda_i$ with respect to $S$ is $u_i u_i^\top$ \cite{magnus1985differentiating}. $\Lnorm$ is not symmetric, but it has the same set of eigenvalues as the symmetric matrix $L_\text{sym} := D^{-1/2}LD^{-1/2}$. So, we can take the derivative of eigenvalues with respect to $L_\text{sym}$, which itself is a differentiable function of the network weights,
allowing us to regularize the eigenvalues of $\Lnorm$. For a derivation of this gradient, see appendix~\ref{app:clust_reg_math}.
To ensure that produced clusterability is meaningful, we also normalize the weights as described in appendix~\ref{app:clust_reg_norm}.

Due to the expense of eigenvalue computation on large networks\footnote{Expense that was largely caused by a coding problem that was only noticed by the time this paper was ready to publish.}, we tested this regularizer on an MLP with 3 hidden layers of width 64 that was trained on MNIST images downsampled to $7 \times 7$ pixels. We trained 10 networks without regularization and 10 with the 2\ts{nd}, 3\ts{rd}, and 4\ts{th} eigenvalues regularized with weight 0.1. As we see in table~\ref{tab:clust_grad}, little accuracy is lost, and regularized networks are more clusterable
both in absolute and relative terms.
Results are also plotted in figure~\ref{fig:clust_reg_scatter}.

\begin{table*}
\fontsize{9pt}{11pt}\selectfont
\centering
\begin{tabular}{ccccccc}
\toprule
Clust.\ reg?      & Pruning? & N-cut    & Dist.\ n-cuts      & Prop.\ $p<0.02$ & Train acc.\ & Test acc.\ \\
\midrule
$\times$          & $\times$ & $10.020$ & $10.035 \pm 0.030$ & $2/10$          & $0.955$     & $0.957$    \\
$\times$          & $\surd$  & $7.11$   & $7.43 \pm 0.12$    & $4/10$          & $0.930$     & $0.934$    \\
$\surd$           & $\times$ & $8.93$   & $8.14 \pm 0.23$    & $1/10$          & $0.961$     & $0.959$    \\
$\surd$           & $\surd$  & $5.51$   & $6.31 \pm 0.20$    & $7/10$          & $0.888$     & $0.896$    \\
\bottomrule
\end{tabular}
\caption{Clusterability of networks trained with and without the clusterability regularizer, with and without pruning. ``Dist.\ n-cuts" contains the mean and standard deviation of the distribution of n-cuts of shuffled networks. All figures shown are averaged over 10 training runs, except ``Prop.\ $p<0.02$", which shows how many networks were more clusterable than all 50 shuffles.}
\label{tab:clust_grad}
\end{table*}

\subsection{Initialization}
\label{subsec:modular_initialization}

Another way to promote clusterability is to initialize weights to be clusterable, and from then on to train
as usual.
After standard initialization, we randomly associate each neuron in the hidden layers with 
one of $c$
tags.
We then weaken the weight of each edge going between differently-tagged neurons by multiplying it by $\beta \in (0,1)$,
and strengthen each edge going between neurons with the same tag by a factor of $1 + (1-\beta)(c-1)$. This preserves the mean absolute value of weights in a layer. This method is computationally cheaper than ongoing regularization, since it only needs to be done once, and doesn't require the computation of eigenvalues.

\begin{figure}
    \centering
    \includegraphics[width=0.9\columnwidth]{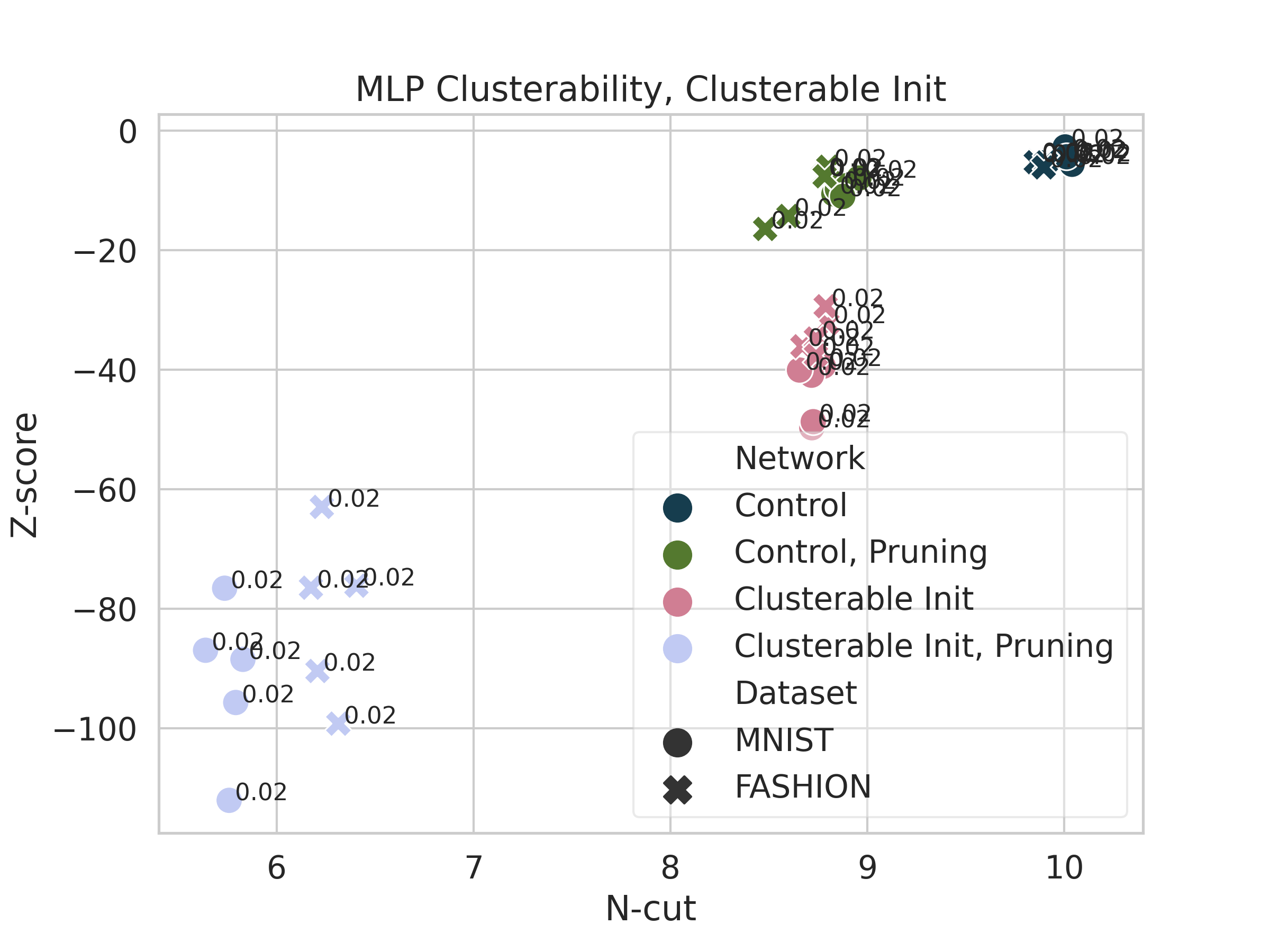}
    \caption{Clusterability of MLPs trained with and without clusterable initialization. Points are labeled with their one-sided $p$-value.}
    \label{fig:mlp_clust_init}
\end{figure}

\begin{figure}
    \centering
    \includegraphics[width=0.9\columnwidth]{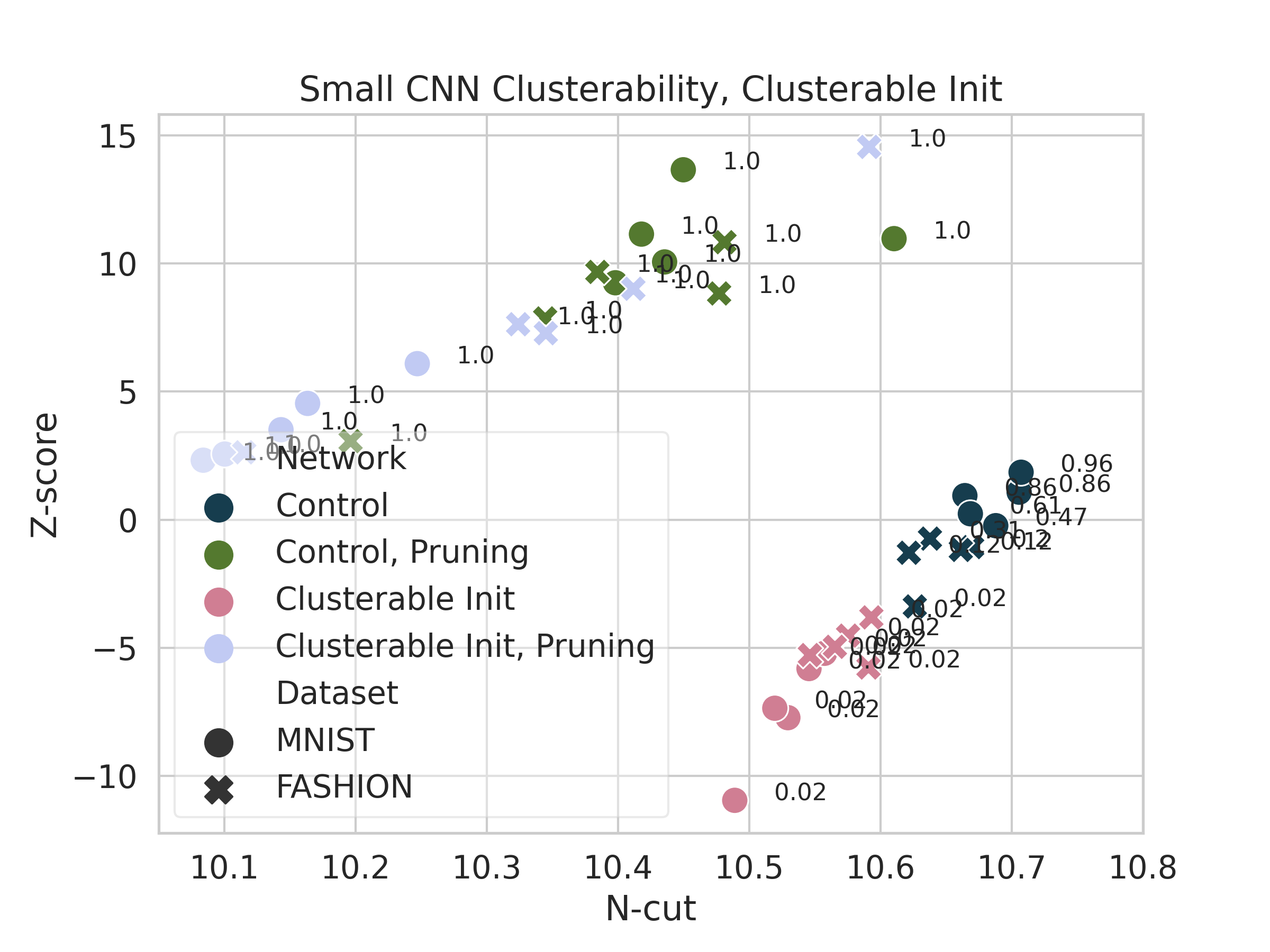}
    \caption{Clusterability of small CNNs trained with and without clusterable initialization. Points are labeled with their one-sided $p$-value.}
    \label{fig:cnn_clust_init}
\end{figure}

\begin{figure}
    \centering
    \includegraphics[width=0.9\columnwidth]{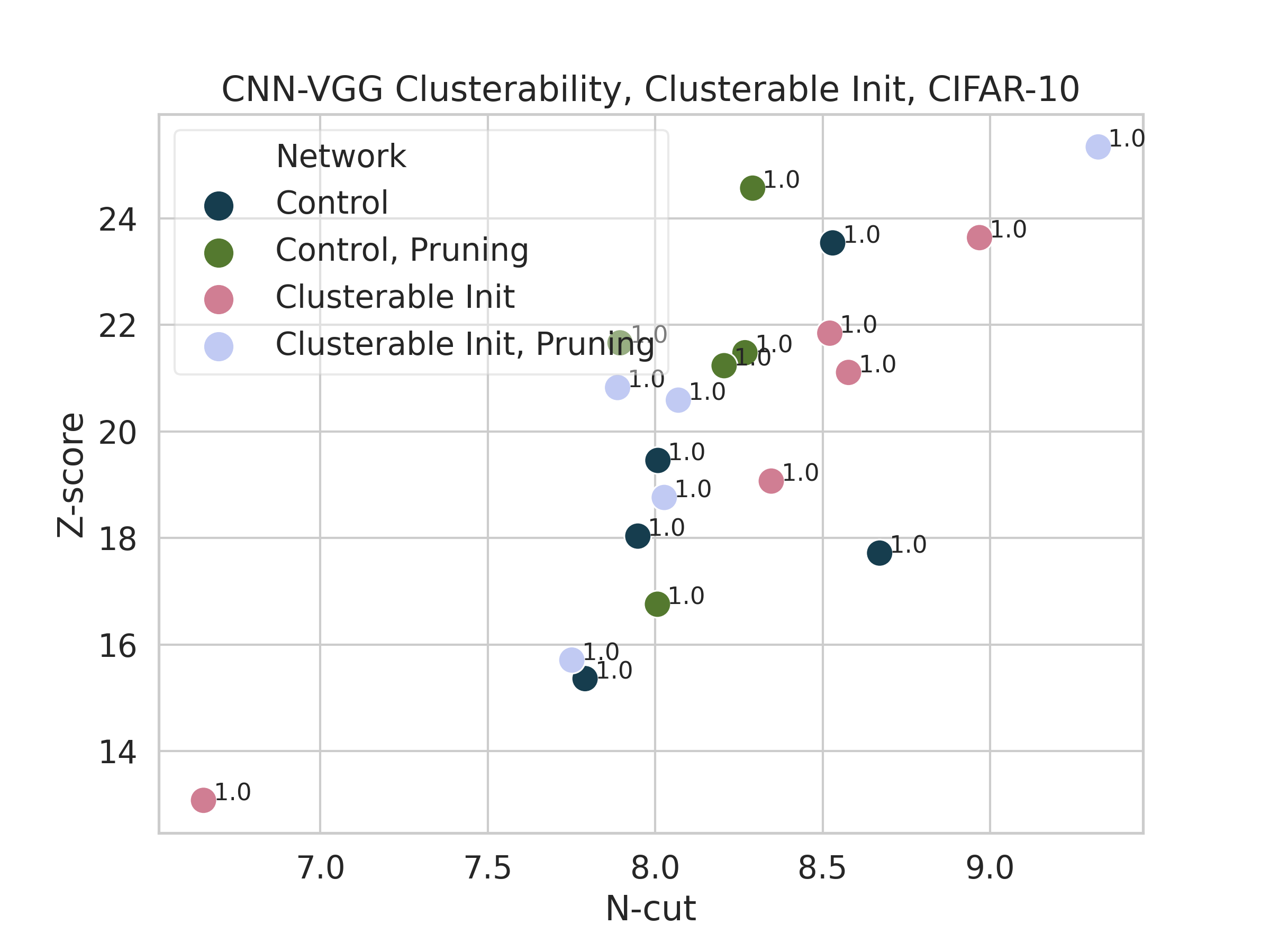}
    \caption{Clusterability of VGGs trained with and without clusterable initialization. Points are labeled with their one-sided $p$-value.}
    \label{fig:vgg_clust_init}
\end{figure}

We trained clusterably initialized MLPs on MNIST and Fashion-MNIST with $c=10$ and $\beta=0.6$, chosen to balance performance with final clusterability. We saw no loss in test accuracy. Results are shown in figure~\ref{fig:mlp_clust_init} and table~\ref{tab:clust_init_mlp_stats}. This initialization method appears to be effective in promoting relative and absolute clusterability, especially in tandem with weight pruning. We also trained clusterably initialized CNNs on MNIST and Fashion-MNIST, and a clusterably initialized VGG on CIFAR-10. For these, we kept $c=10$ but instead used $\beta=0.8$. Again, we saw no loss in test accuracy. Results are shown in figures~\ref{fig:cnn_clust_init} and \ref{fig:vgg_clust_init} and tables~\ref{tab:clust_init_small_cnn_stats} and \ref{tab:clust_init_vgg_stats}. By contrast with the MLP results, the initialization seems much less effective at promoting absolute and relative clusterability.

\section{Related Work}
\label{sec:related_work}

Previous research has also explored learned modularity in neural networks. Watanabe et al.\ \shortcite{watanabe2018modular, watanabe2018understanding, watanabe2019interpreting} demonstrate a different way of grouping neurons into modules, investigate properties of the modules, and link their notion of modularity to generalization error. Davis et al.\ \shortcite{davis2019nif} use a clustering method based on statistical properties (rather than the weights of the network directly) to visualize networks, prune them, and do feature attribution. Lu and Ester \shortcite{lu2019checking} bi-cluster neurons in a hidden layer and show that this divides neurons into functionally-related groups that are important for classification. 
None of these papers compare the modularity of networks trained in different ways, nor do they show that networks are more modular than initialized networks or networks with the same per-layer distribution of weights.
There is also existing literature on
imposing modular structure on neural networks,
either by some kind of dynamic routing \cite{andreas2016neural, chang2018automatically, kirsch2018modular, alet2018modular} or by explicitly enforcing the desired modularity structure \cite{lee2018modular, oyama2001inverse}. These typically involve training with a non-standard architecture, limiting their broader applicability. Amer and Maul \shortcite{amer2019review} offer a review of modularization techniques.

A number of papers have investigated some aspect of the structure of neural networks. Similar contributions to this work include those by Cammarata et al.\ \shortcite{cammarata2020thread}, who investigate small groups of neurons that evaluate some intelligible function inside Inception-V1, Frankle and Carbin \shortcite{frankle2018the}, who discover that trained neural networks contain efficiently-trainable sub-networks, inspiring multiple follow-up papers \cite{zhou2019deconstructing, frankle2019linear}, and You et al.\ \cite{you2020graph} who form a `relational graph' from a neural network, and study how network performance relates to properties of the relational graph.

\section{Conclusion}
\label{sec:conclusion}

We see that many neural networks are clusterable to a degree that would not be expected merely from their initialization, or from their distributions of weights.
MLPs trained on image classification are reliably more clusterable than 98\% of their shuffles when $L_1$ or $L_2$ regularization is not used, as are networks trained on ImageNet classification. So are VGGs trained on CIFAR-10, as long as they are trained with dropout (which is standard practice) and are pruned (which is not).
$L_1$ and $L_2$ regularization appear to have irregular effects on clusterability, sometimes producing spuriously clusterable networks, and sometimes producing remarkably unclusterable networks.
We also demonstrate that clusterability can be induced in MLPs either by initializing weights differently or by regularization, with little cost to performance and without hand-coding a non-standard architecture or significantly modifying the training procedure.

Overall, we find that clusterability is a common phenomenon, giving hope that networks are also in some sense modular.
That being said, our work limits itself to image classification networks, leaving open the question of clusterability in other domains.
We also fail to induce clusterability in CNNs by initialization or regularization, which suggests a direction for future work.

\section*{Acknowledgments}
\label{sec:ack}

The authors would like to thank Open Philanthropy for their financial support of this research, the researchers at UC Berkeley's Center for Human-Compatible AI for their advice, and anonymous reviewers for their contributions to improving the paper. Daniel Filan would like to thank Paul Christiano, Rohin Shah, Matthew `Vaniver' Graves, and Buck Shlegeris for valuable discussions that helped shape this research, as well as Andrei Knyazev for his work in helping debug scikit-learn's implementation of spectral clustering. Shlomi Hod would like to thank Dmitrii Krasheninnikov for fruitful conversations throughout the summer of 2019.

\bibliography{main}

\input{supplementary.tex}

\end{document}

%% file: notation.tex
\newcommand{\ncut}{\textrm{n-cut}}
\newcommand{\vol}{\textrm{vol}}
\newcommand{\Lnorm}{L_\textrm{norm}}
\newcommand{\Lsym}{L_\textrm{sym}}
\newcommand{\R}{\mathbb{R}}
\newcommand{\sgn}{\textrm{sgn}}

\newcommand{\ts}[1]{\textsuperscript{#1}}

%% file: supplementary.tex
\onecolumn
\appendix

\numberwithin{table}{section}
\numberwithin{figure}{section}

\section{Supplementary Material}

\subsection{Probabilistic Interpretation of N-cut}
\label{app:ncut_meaning}

\input{appendix_sections/ncut_meaning}

\subsection{Training Details}
\label{app:train_hypers}

\input{appendix_sections/training_hyperparams}

\subsection{Choosing the Number of Clusters}
\label{app:n_clusters}

\input{appendix_sections/n_clusters}

\subsection{Clusterability Data}
\label{app:clusterability_data}

\input{appendix_sections/clust_data}

\subsection{Random Dataset Experiments}
\label{app:random_dataset_experiments}

\input{appendix_sections/random_datasets}

\subsection{Topology-Preserving Shuffles}
\label{app:topology_preserving_shuffles}

\input{appendix_sections/topology-preserving-shuffles}

\subsection{Polynomial Regression}
\label{app:poly_regression}

\input{appendix_sections/poly_regression}

\subsection{Mixture Dataset Results}
\label{app:mixture_datasets}

\input{appendix_sections/mixture_results}

\subsection{Derivation of Clusterability Regularizer}
\label{app:clust_reg_math}

\input{appendix_sections/clust_grad}

\subsection{Normalization for Clusterability Regularization}
\label{app:clust_reg_norm}

\input{appendix_sections/clust_reg_norm}

%% file: appendix_sections/ncut_meaning.tex
As well as the formal definition given in section \ref{sec:defs_and_background}, n-cut has a more intuitive interpretation. Note that a similar argument appears in \citet{meila2001random}.

Divide each edge between vertices $i$ and $j$ into two `stubs', one attached to $i$ and the other attached to $j$, and associate with each stub the weight of the whole edge. Now: suppose $(X_1, \dotsc, X_k)$ is a partition of the graph. First, pick an integer $l$ between 1 and $k$ uniformly at random. Secondly, out of all of the stubs attached to vertices in $X_l$, pick one with probability proportional to its weight. Say that this procedure `succeeds' if the edge associated with that stub connects two vertices inside $X_l$, and `fails' if the edge connects a vertex inside $X_l$ with a vertex outside $X_l$. The probability that the procedure fails is $\ncut(X_1, \dotsc, X_k)/k$.

Therefore, the n-cut divided by $k$ is roughly a measure of what proportion of edge weight coming from vertices inside a partition element crosses the partition boundary.

%% file: appendix_sections/training_hyperparams.tex
During training, we use the Adam algorithm \citep{kingma2014adam} with the standard Keras hyperparameters: learning rate $0.001$, $\beta_1 = 0.9$, $\beta_2 = 0.999$, no amsgrad. The loss function was categorical cross-entropy except for networks trained on polynomial regression (see appendix~\ref{app:poly_regression}, where we used mean squared error). For pruning, our initial sparsity is 0.5, our final sparsity is 0.9, the pruning frequency is 10 steps, and we use a cubic pruning schedule (see \citet{zhu2017prune}). Initial and final sparsities were chosen due to their use in the TensorFlow Model Optimization Tutorial.\footnote{URL: \url{https://web.archive.org/web/20190817115045/https://www.tensorflow.org/model_optimization/guide/pruning/pruning_with_keras}}
A batch size of 128 was used for the MLPs and VGGs, while a batch size of 64 was used for the small CNNs.
For MLPs, training went for 20 epochs before pruning and 20 epochs of pruning. For small CNNs, it was 10 epochs before pruning and 10 epochs of pruning. For VGGs trained on CIFAR-10, it was 200 epochs before pruning and 50 epochs of pruning. For networks trained on $7 \times 7$ MNIST to test clusterability regularization, it was 10 epochs before pruning and 10 epochs of pruning.
The small CNNs trained on MNIST, Fashion-MNIST, and the stack datasets have all convolutional kernels being 3 by 3, with the second and third hidden layers being followed by max pooling with a 2 by 2 window.
For training the VGG on CIFAR-10, data augmentations used are random rotations between 0 and 15 degrees, random shifts both vertically and horizontally of up to 10\% of the side length, and random horizontal flipping.
We use Tensorflow's implementation of the Keras API \cite{tensorflow2015-whitepaper, chollet2015keras}.

%% file: appendix_sections/n_clusters.tex
The number of clusters is a hyperparameter of the spectral clustering algorithm. In the paper we reported the results from using 12 clusters.

To test the robustness of our results to this hyperparameter, we re-ran the MLP clusterability experiments on pruned networks trained with and without dropout, using 2, 4, 7, and 10 clusters. In each condition, we trained 10 networks, compared to a distribution of 320 shuffles, and counted how many networks were more clusterable than all 320 of the shuffles.
For 4, 7, and 10 clusters, the results aligned with what we reported in section~\ref{sec:mlp_clusterability}: networks were relatively clusterable, and more clusterable (both absolutely and relatively) when trained with dropout.
including stronger significance patterns and lower mean n-cut for dropout-trained models vs. without dropout. Results are shown in tables~\ref{tab:ncut-stats-4-clusters}, \ref{tab:ncut-stats-7-clusters}, and \ref{tab:ncut-stats-10-clusters} respectively.

\begin{table}[h!]
\centering
\begin{tabular}{lcccccc}
\toprule
Dataset & Dropout & N-cuts & Dist.\ n-cuts & Prop.\ $p < 1/320$ & Train acc.\ & Test acc \\
\midrule
MNIST & $\times$ & $2.000 \pm 0.035$ & $2.042 \pm 0.017$ & $7/10$ & $1.00$ & $0.984$ \\
MNIST & $\surd$ & $1.840 \pm 0.015$ & $2.039 \pm 0.019$ & $10/10$ & $0.967$ & $0.979$ \\
Fashion & $\times$ & $1.880 \pm 0.030$ & $1.992 \pm 0.018$ & $10/10$ & $0.983$ & $0.893$ \\
Fashion & $\surd$ & $1.726 \pm 0.022$ & $2.013 \pm 0.017$ & $10/10$ & $0.863$ & $0.869$ \\
\bottomrule
\end{tabular}
\caption{Clusterability results for pruned MLPs with 4 clusters. N-cuts shows the mean and standard deviations of the n-cuts of the trained networks. Dist.\ n-cuts shows the average mean and standard deviation of the distributions of shuffles, the average being taken over the 10 trained networks. Prop.\ $p < 1/320$ shows how many networks were more clusterable than all 320 shuffles. Accuracies are averaged over runs.}
\label{tab:ncut-stats-4-clusters}
\end{table}

\begin{table}
\centering
\begin{tabular}{lccccccc}
\toprule
Dataset & Dropout & N-cuts & Dist.\ n-cuts & Prop.\ $p < 1/320$ & Train acc.\ & Test acc.\ \\
\midrule
MNIST    & $\times$  & $4.556 \pm 0.052$ & $4.710 \pm 0.026$ & $10/10$ & $1.00 $ & $0.984$  \\
MNIST    & $\surd$   & $4.222 \pm 0.035$ & $4.712 \pm 0.023$ & $10/10$ & $0.967$ & $0.979$  \\
Fashion  & $\times$  & $4.351 \pm 0.057$ & $4.613 \pm 0.029$ & $10/10$ & $0.983$ & $0.893$  \\
Fashion  & $\surd$   & $4.079 \pm 0.046$ & $4.663 \pm 0.029$ & $10/10$ & $0.863$ & $0.869$  \\
\bottomrule
\end{tabular}
\caption{
Clusterability results for pruned MLPs with 7 clusters. Reporting as in table~\ref{tab:ncut-stats-4-clusters}.
}
\label{tab:ncut-stats-7-clusters}
\end{table}

\begin{table}
\centering
\begin{tabular}{lccccccc}
\toprule
Dataset & Dropout &  N-cuts & Dist.\ n-cuts & Prop.\ $p < 1/320$ & Train acc.\ & Test acc.\ \\
\midrule
MNIST    & $\times$ & $7.137 \pm 0.042$ & $7.397 \pm 0.037$ & $10/10$ & $1.00 $ & $0.984$  \\
MNIST    & $\surd$  & $6.688 \pm 0.035$ & $7.421 \pm 0.031$ & $10/10$ & $0.967$ & $0.979$  \\
Fashion  & $\times$ & $6.975 \pm 0.117$ & $7.257 \pm 0.036$ & $ 9/10$ & $0.983$ & $0.893$  \\
Fashion  & $\surd$  & $6.460 \pm 0.041$ & $7.339 \pm 0.034$ & $10/10$ & $0.863$ & $0.869$  \\
\bottomrule
\end{tabular}
\caption{
Clusterability results for pruned MLPs with 10 clusters. Reporting as in table~\ref{tab:ncut-stats-4-clusters}
}
\label{tab:ncut-stats-10-clusters}
\end{table}

When we ran the experiments with 2 clusters, the significance results were very different, as shown in table~\ref{tab:ncut-stats-2-clusters}. No network trained on MNIST or Fashion-MNIST was statistically significantly clusterable. As such, we conjecture that our results generalize to any number of clusters that is not too small (or comparable to the number of neurons).

\begin{table}
\centering
\begin{tabular}{lccccccc}
\toprule
Dataset & Dropout & N-cuts & Dist.\ n-cuts & Prop.\ $p < 1/320$ & Train acc.\ & Test acc.\ \\
\midrule
MNIST    & $\times$ & $0.333 \pm 0.003$ & $0.330 \pm 0.003$ & $0/10$ & $1.00 $ & $0.984$  \\
MNIST    & $\surd$  & $0.332 \pm 0.002$ & $0.323 \pm 0.003$ & $0/10$ & $0.967$ & $0.979$  \\
Fashion  & $\times$ & $0.319 \pm 0.003$ & $0.313 \pm 0.003$ & $0/10$ & $0.983$ & $0.893$  \\
Fashion  & $\surd$  & $0.312 \pm 0.003$ & $0.312 \pm 0.003$ & $0/10$ & $0.863$ & $0.869$  \\
\bottomrule
\end{tabular}
\caption{
Clusterability results for pruned MLPs with 2 clusters. Reporting as in table~\ref{tab:ncut-stats-4-clusters}.
}
\label{tab:ncut-stats-2-clusters}
\end{table}

%% file: appendix_sections/clust_data.tex
In this section, we give tables and plots providing more information on data that was presented in the paper. 

\begin{figure}
    \centering
    \includegraphics[width=0.5\columnwidth]{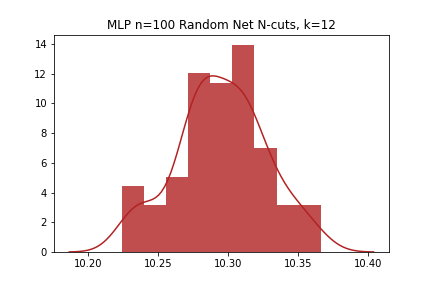}
    \caption{N-cuts of 100 randomly-initialized MLPs. Vertical axis shows probability density. Plot was generated using the \texttt{distplot} function of seaborn 0.9.0 \cite{seaborn_0_9_0} with \texttt{kde} set to \texttt{True}.}
    \label{fig:mlp_init_ncuts}
\end{figure}

\begin{figure}
    \centering
    \includegraphics[width=0.5\columnwidth]{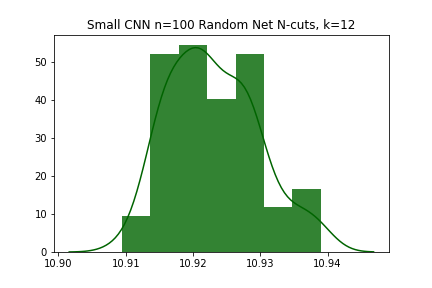}
    \caption{N-cuts of 100 randomly-initialized small CNNs. Vertical axis shows probability density. Plot was generated using the \texttt{distplot} function of seaborn 0.9.0 \cite{seaborn_0_9_0} with \texttt{kde} set to \texttt{True}.}
    \label{fig:small_cnn_init_ncuts}
\end{figure}

\begin{figure}
    \centering
    \includegraphics[width=0.5\columnwidth]{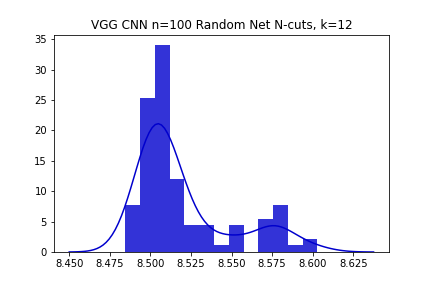}
    \caption{N-cuts of 100 randomly-initialized VGGs. Vertical axis shows probability density. Plot was generated using the \texttt{distplot} function of seaborn 0.9.0 \cite{seaborn_0_9_0} with \texttt{kde} set to \texttt{True}.}
    \label{fig:vgg_init_ncuts}
\end{figure}

First, figures~\ref{fig:mlp_init_ncuts}, \ref{fig:small_cnn_init_ncuts}, and \ref{fig:vgg_init_ncuts} show the distribution of n-cuts of randomly-initialized MLPs, small CNNs, and VGGs, respectively.

We also show tables of statistics of clusterability experiments shown in the main paper: tables~\ref{tab:mlp_mnist_stats}, \ref{tab:mlp_fashion_stats}, \ref{tab:mlp_halves_stats}, \ref{tab:small_cnn_mnist_stats}, \ref{tab:small_cnn_fashion_stats}, \ref{tab:cifar_vgg_stats}, \ref{tab:imagenet_table}, \ref{tab:small_cnn_stack_stats}, \ref{tab:clust_init_mlp_stats}, \ref{tab:clust_init_small_cnn_stats}, and \ref{tab:clust_init_vgg_stats}. In these, the column ``N-cut" shows the mean n-cut of the networks, ``Dist.\ n-cuts" shows the average mean and standard deviation of the distribution of n-cuts of shuffled networks, where the average is taken over different trained networks (which induce different distributions of shuffled networks), and train and test accuracies are also averaged over runs. In table~\ref{tab:imagenet_table}, since we only have one of each network, no averaging is done. ``Prop.\ $p < 0.02$" shows how many networks are more clusterable than all 50 shuffles.

\begin{figure}
    \centering
    \includegraphics[width=0.7\textwidth]{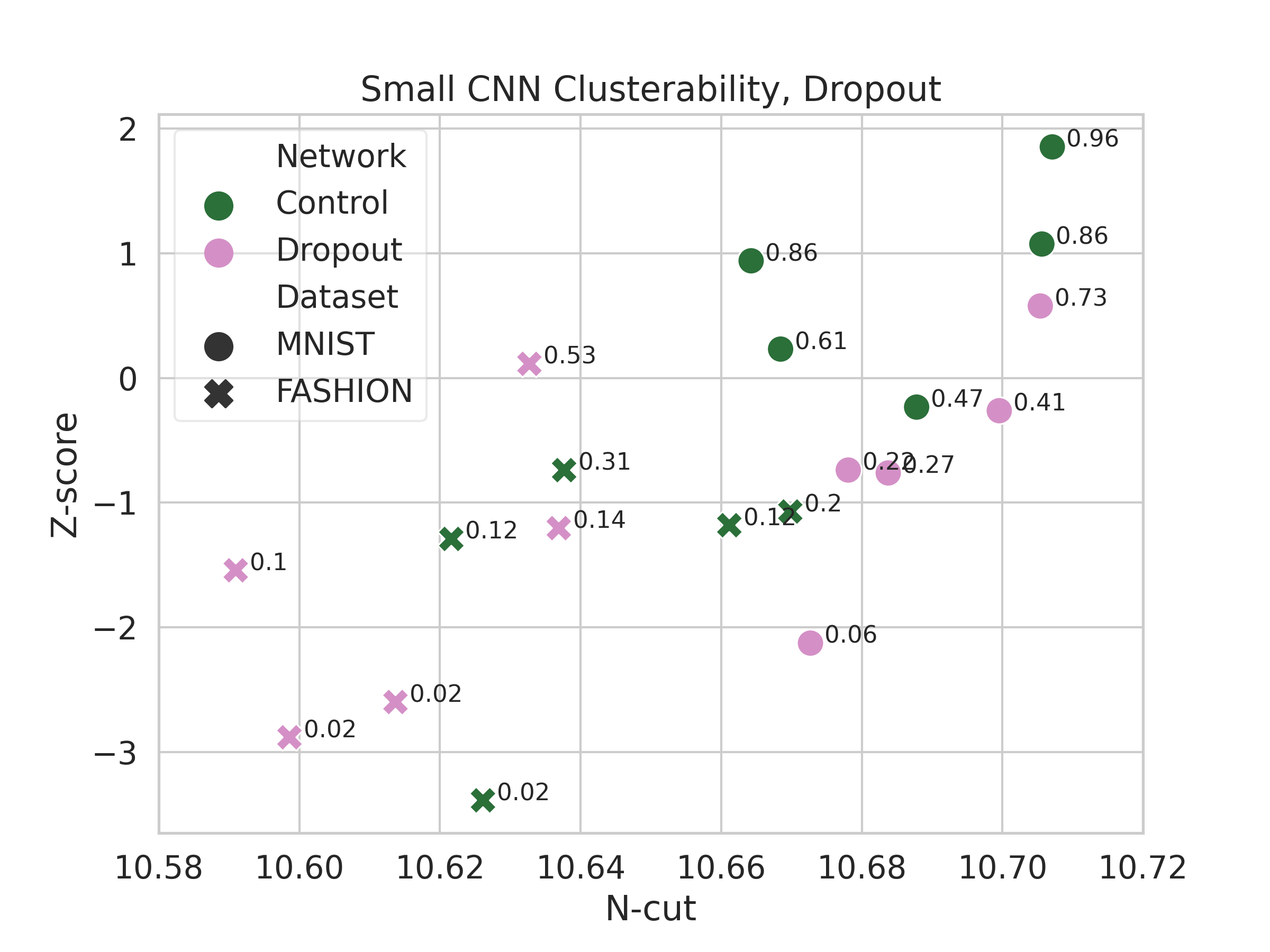}
    \caption{Clusterability of small CNNs trained without pruning with and without dropout---a subset of figure~\ref{fig:small_cnn_unpruned}.}
    \label{fig:small_cnn_unpruned_focus}
\end{figure}

Figure~\ref{fig:small_cnn_unpruned_focus} plots the clusterability of unpruned small CNNs trained with and without dropout. This serves to zoom in on a region of figure~\ref{fig:small_cnn_unpruned}, which also plots $L_1$-{} and $L_2$-regularized networks, where it's hard to make out the details of the networks trained with no regularization or dropout.

\begin{figure}
\centering
\includegraphics[width=0.7\textwidth]{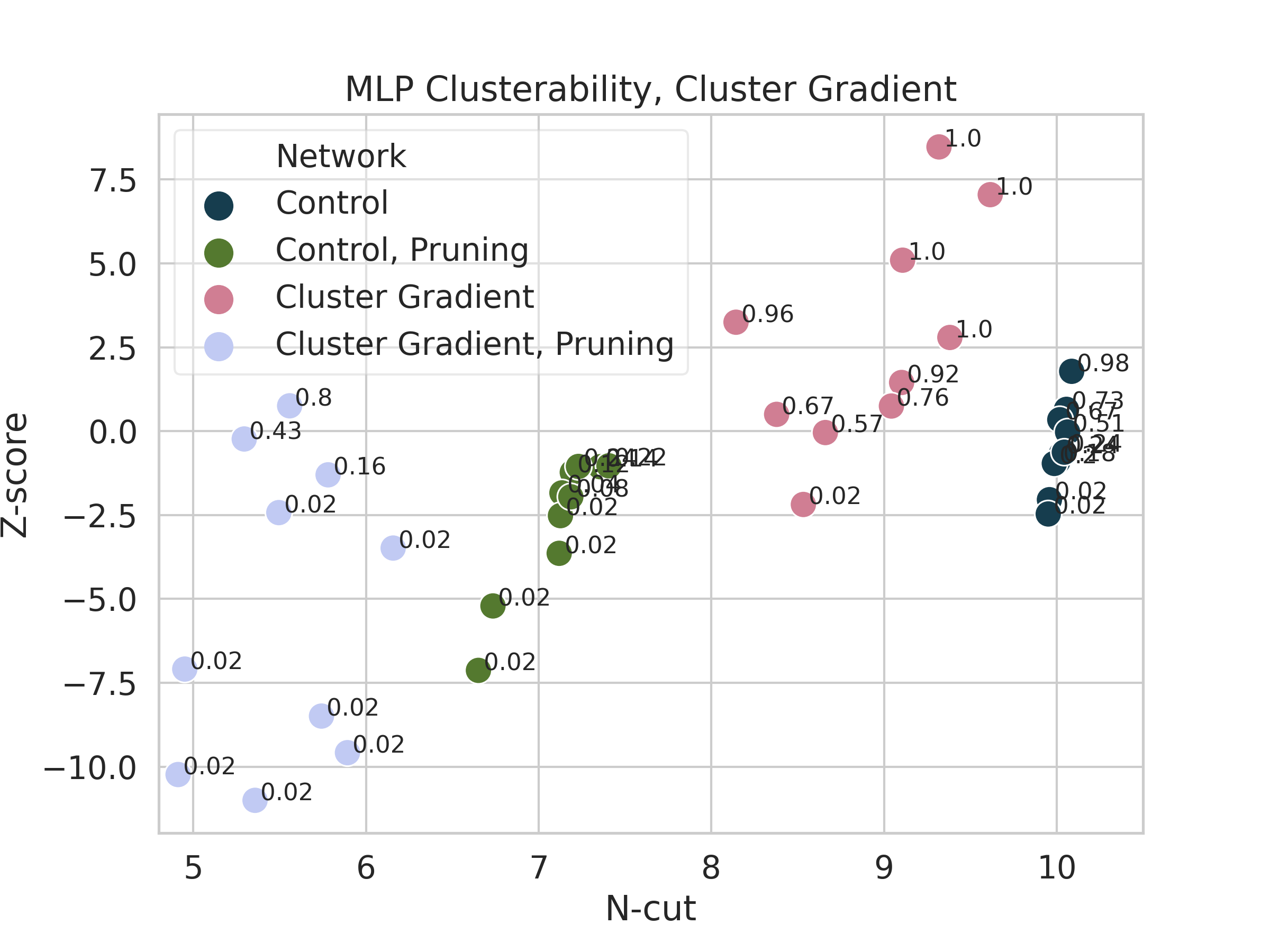}
\caption{Clusterability of small MLPs trained with and without the clusterability regularizer. Points are labeled with their one-sided $p$-value.}
\label{fig:clust_reg_scatter}
\end{figure}

Finally, figure~\ref{fig:clust_reg_scatter} shows a scatter plot of the clusterability of small MLPs trained with and without clusterability regularization.

\begin{table}
\centering
\begin{tabular}{lcccccc}
\toprule
Reg.\ method & Pruning? & N-cut    & Dist.\ n-cuts      & Prop.\ $p<0.02$ & Train acc.\ & Test acc.\ \\
\midrule
None         & $\times$ & $10.024$ & $10.165 \pm 0.035$ & $5/5$           & $0.997$     & $0.980$    \\
Dropout      & $\times$ & $9.97$   & $10.30 \pm 0.11$   & $5/5$           & $0.973$     & $0.980$    \\
$L_1$        & $\times$ & $9.06$   & $6.50 \pm 0.14$    & $0/5$           & $0.991$     & $0.979$    \\
$L_2$        & $\times$ & $9.639$  & $9.387 \pm 0.066$  & $1/5$           & $0.992$     & $0.980$    \\
None         & $\surd$  & $8.880$  & $9.183 \pm 0.031$  & $5/5$           & $1.000$     & $0.984$    \\
Dropout      & $\surd$  & $8.350$  & $9.247 \pm 0.039$  & $5/5$           & $0.967$     & $0.979$    \\
$L_1$        & $\surd$  & $10.063$ & $2.72 \pm 0.33$    & $0/5$           & $0.997$     & $0.981$    \\
$L_2$        & $\surd$  & $8.335$  & $8.599 \pm 0.081$  & $3/5$           & $0.998$     & $0.982$    \\
\bottomrule
\end{tabular}
\caption{Clusterability of MLPs trained on MNIST. See start of appendix~\ref{app:clusterability_data} for details.}
\label{tab:mlp_mnist_stats}
\end{table}

\begin{table}
\centering
\begin{tabular}{lcccccc}
\toprule
Reg.\ method & Pruning? & N-cut   & Dist.\ n-cuts      & Prop.\ $p<0.02$ & Train acc.\ & Test acc.\ \\
\midrule
None         & $\times$ & $9.884$ & $10.080 \pm 0.037$ & $5/5$           & $0.939$     & $0.891$    \\
Dropout      & $\times$ & $9.698$ & $10.052 \pm 0.034$ & $5/5$           & $0.862$     & $0.872$    \\
$L_1$        & $\times$ & $8.41$  & $7.21 \pm 0.10$    & $0/5$           & $0.907$     & $0.880$    \\
$L_2$        & $\times$ & $9.468$ & $9.185 \pm 0.068$  & $0/5$           & $0.913$     & $0.879$    \\
None         & $\surd$  & $8.687$ & $9.032 \pm 0.035$  & $5/5$           & $0.983$     & $0.894$    \\
Dropout      & $\surd$  & $8.110$ & $9.131 \pm 0.041$  & $5/5$           & $0.864$     & $0.867$    \\
$L_1$        & $\surd$  & $9.94$  & $4.59 \pm 0.23$    & $0/5$           & $0.933$     & $0.887$    \\
$L_2$        & $\surd$  & $8.524$ & $8.746 \pm 0.077$  & $3/5$           & $0.951$     & $0.892$    \\
\bottomrule
\end{tabular}
\caption{Clusterability of MLPs trained on Fashion-MNIST. See start of appendix~\ref{app:clusterability_data} for details.}
\label{tab:mlp_fashion_stats}
\end{table}

\begin{table}
\centering
\begin{tabular}{cccccccc}
\toprule
Dataset & Same or diff & Pruning? & N-cut    & Dist.\ n-cuts      & Prop.\ $p<0.02$ & Train acc.\ & Test acc.\ \\
\midrule
MNIST   & same         & $\times$ & $10.041$ & $10.104 \pm 0.027$ & $2/5$           & $0.999$     & $0.993$    \\
MNIST   & diff         & $\times$ & $9.874$  & $10.312 \pm 0.091$ & $5/5$           & $0.989$     & $0.926$    \\
MNIST   & same         & $\surd$  & $8.923$  & $9.036 \pm 0.035$  & $3/5$           & $1.000$     & $0.995$    \\
MNIST   & diff         & $\surd$  & $8.460$  & $9.181 \pm 0.033$  & $5/5$           & $1.000$     & $0.940$    \\
Fashion & same         & $\times$ & $9.782$  & $9.919 \pm 0.032$  & $4/5$           & $0.972$     & $0.940$    \\
Fashion & diff         & $\times$ & $9.881$  & $10.246 \pm 0.021$ & $5/5$           & $0.849$     & $0.716$    \\
Fashion & same         & $\surd$  & $8.447$  & $8.800 \pm 0.038$  & $4/5$           & $0.997$     & $0.947$    \\
Fashion & diff         & $\surd$  & $8.637$  & $9.046 \pm 0.028$  & $5/5$           & $0.899$     & $0.724$    \\
\bottomrule
\end{tabular}
\caption{Clusterability of MLPs trained on `halves' datasets. See start of appendix~\ref{app:clusterability_data} for details.}
\label{tab:mlp_halves_stats}
\end{table}

\begin{table}
\centering
\begin{tabular}{lcccccc}
\toprule
Reg.\ method & Pruning? & N-cut    & Dist.\ n-cuts      & Prop.\ $p<0.02$ & Train acc.\ & Test acc.\ \\
\midrule
None         & $\times$ & $10.687$ & $10.673 \pm 0.018$ & $0/5$           & $0.998$     & $0.992$    \\
Dropout      & $\times$ & $10.688$ & $10.698 \pm 0.016$ & $0/5$           & $0.990$     & $0.994$    \\
$L_1$        & $\times$ & $10.31$  & $5.30 \pm 0.18$    & $0/5$           & $0.993$     & $0.991$    \\
$L_2$        & $\times$ & $8.050$  & $10.252 \pm 0.032$ & $5/5$           & $0.996$     & $0.991$    \\
None         & $\surd$  & $10.462$ & $10.034 \pm 0.039$ & $0/5$           & $1.000$     & $0.993$    \\
Dropout      & $\surd$  & $10.267$ & $10.073 \pm 0.039$ & $0/5$           & $0.974$     & $0.993$    \\
$L_1$        & $\surd$  & $9.31$   & $3.05 \pm 0.28$    & $0/5$           & $0.995$     & $0.991$    \\
$L_2$        & $\surd$  & $10.754$ & $9.764 \pm 0.059$  & $0/5$           & $0.999$     & $0.992$    \\
\bottomrule
\end{tabular}
\caption{Clusterability of small CNNs trained on MNIST. See start of appendix~\ref{app:clusterability_data} for details.}
\label{tab:small_cnn_mnist_stats}
\end{table}

\begin{table}
\centering
\begin{tabular}{lcccccc}
\toprule
Reg.\ method & Pruning? & N-cut    & Dist.\ n-cuts      & Prop.\ $p<0.02$ & Train acc.\ & Test acc.\ \\
\midrule
None         & $\times$ & $10.643$ & $10.671 \pm 0.018$ & $1/5$           & $0.982$     & $0.922$    \\
Dropout      & $\times$ & $10.615$ & $10.646 \pm 0.019$ & $2/5$           & $0.922$     & $0.925$    \\
$L_1$        & $\times$ & $9.98$   & $7.49 \pm 0.11$    & $0/5$           & $0.939$     & $0.913$    \\
$L_2$        & $\times$ & $9.303$  & $10.578 \pm 0.023$ & $5/5$           & $0.972$     & $0.923$    \\
None         & $\surd$  & $10.377$ & $10.060 \pm 0.041$ & $0/5$           & $0.995$     & $0.922$    \\
Dropout      & $\surd$  & $9.974$  & $9.951 \pm 0.042$  & $0/5$           & $0.863$     & $0.905$    \\
$L_1$        & $\surd$  & $10.28$  & $8.62 \pm 0.13$    & $0/5$           & $0.954$     & $0.919$    \\
$L_2$        & $\surd$  & $10.445$ & $9.976 \pm 0.043$  & $0/5$           & $0.994$     & $0.923$    \\
\bottomrule
\end{tabular}
\caption{Clusterability of small CNNs trained on Fashion-MNIST. See start of appendix~\ref{app:clusterability_data} for details.}
\label{tab:small_cnn_fashion_stats}
\end{table}

\begin{table}
\centering
\begin{tabular}{lcccccc}
\toprule
Reg.\ method & Pruning? &  N-cut &  Dist.\ n-cuts & Prop.\ $p<0.02$ & Train acc.\ & Test acc.\  \\
\midrule
None         & $\times$   &  $8.189$ &          $3.819 \pm 0.235$ &    $ 0/5$ &    $0.958$ &   $0.882$ \\
Dropout         & $\times$   &  $8.844$ &          $8.283 \pm 0.117$ &    $ 0/5$ &    $0.904$ &   $0.896$ \\
$L_1$          & $\times$   &  $5.294$ &          $2.074 \pm 0.036$ &    $ 0/5$ &    $0.912$ &   $0.871$ \\
$L_2$          & $\times$   &  $8.548$ &          $2.104 \pm 0.194$ &    $ 0/5$ &    $0.958$ &   $0.899$ \\
Dropout \& $L_2$ & $\times$   &  $8.355$ &          $8.246 \pm 0.101$ &    $ 1/5$ &    $0.900$ &   $0.898$ \\
Control  & $\surd$   &  $8.133$ &          $3.603 \pm 0.213$ &    $ 0/5$ &    $0.974$ &   $0.909$ \\
Dropout  & $\surd$   &  $6.341$ &          $7.977 \pm 0.206$ &    $ 5/5$ &    $0.905$ &   $0.905$ \\
$L_1$   & $\surd$   &  $4.701$ &          $0.741 \pm 0.052$ &    $ 0/5$ &    $0.930$ &   $0.881$ \\
$L_2$   & $\surd$   &  $8.701$ &          $1.730 \pm 0.223$ &    $ 0/5$ &    $0.974$ &   $0.908$ \\
Dropout \& $L_2$ & $\surd$   &  $6.135$ &          $7.994 \pm 0.184$ &    $ 4/5$ &    $0.905$ &   $0.902$ \\
\bottomrule
\end{tabular}
\caption{Clusterability of VGGs trained on CIFAR-10. See start of appendix~\ref{app:clusterability_data} for details.}
\label{tab:cifar_vgg_stats}
\end{table}

\begin{table}
\centering
\begin{tabular}{lcccc}
\toprule
Network      & N-cut   & Dist.\ n-cuts     & $p$-value & Top-1 acc. \\
\midrule
VGG-16       & $7.800$  & $8.213 \pm 0.019$   & $0.01$   & $0.708$    \\
VGG-19       & $7.330$  & $7.655 \pm 0.043$   & $0.01$   & $0.709$    \\
ResNet-18    & $7.90$   & $8.51 \pm 0.28$     & $0.01$   & $0.682$    \\
ResNet-34    & $6.318$  & $6.375 \pm 0.001$   & $0.01$   & $0.722$    \\
ResNet-50    & $4.125$  & $4.185 \pm 0.001$   & $0.01$   & $0.748$    \\
Inception-V3 & $2.906$  & $2.938 \pm 0.007$   & $0.02$   & $0.776$    \\
\bottomrule
\end{tabular}
\caption{Clusterability statistics of networks trained on ImageNet. Top-1 accuracies are taken from \url{https://github.com/qubvel/classification_models}.}
\label{tab:imagenet_table}
\end{table}

\begin{table}
\centering
\begin{tabular}{cccccccc}
\toprule
Dataset & Same or diff & Pruning? &  N-cut   &  Dist.\ n-cuts     &  Prop.\ $p<0.02$ &  Train acc.\ &  Test acc.\   \\
\midrule
MNIST   & same         & $\times$ & $10.702$ & $10.679 \pm 0.017$ &          $ 0/5$  &      $0.999$ &      $0.998$ \\
MNIST   & diff         & $\times$ & $10.548$ & $10.583 \pm 0.021$ &          $ 1/5$  &      $0.979$ &      $0.914$ \\
MNIST   & same         & $\surd$  & $10.347$ & $ 9.969 \pm 0.040$ &          $ 0/5$  &      $1.000$ &      $0.999$ \\
MNIST   & diff         & $\surd$  & $10.225$ & $10.019 \pm 0.039$ &          $ 0/5$  &      $0.978$ &      $0.902$ \\
Fashion & same         & $\times$ & $10.701$ & $10.703 \pm 0.017$ &          $ 0/5$  &      $0.992$ &      $0.958$ \\
Fashion & diff         & $\times$ & $10.524$ & $10.567 \pm 0.020$ &          $ 3/5$  &      $0.860$ &      $0.703$ \\
Fashion & same         & $\surd$  & $10.328$ & $10.109 \pm 0.039$ &          $ 0/5$  &      $1.000$ &      $0.966$ \\
Fashion & diff         & $\surd$  & $ 9.953$ & $ 9.990 \pm 0.044$ &          $ 0/5$  &      $0.818$ &      $0.683$ \\
\bottomrule
\end{tabular}
\caption{Clusterability of small CNNs trained on stack datasets. See start of appendix~\ref{app:clusterability_data} for details.}
\label{tab:small_cnn_stack_stats}
\end{table}

\begin{table}
\centering
\begin{tabular}{lcccccc}
\toprule
Dataset & Pruning? &  N-cut  &  Dist.\ n-cuts      &  Prop.\ $p<0.02$ &  Train acc.\ &  Test acc.\ \\
\midrule
MNIST   & $\times$ & $8.718$ &  $10.065 \pm 0.031$ &            $5/5$ &      $0.996$ &     $0.979$ \\
MNIST   & $\surd$  & $5.750$ &  $ 9.200 \pm 0.038$ &            $5/5$ &      $1.000$ &     $0.983$ \\
Fashion & $\times$ & $8.746$ &  $10.022 \pm 0.038$ &            $5/5$ &      $0.938$ &     $0.892$ \\
Fashion & $\surd$  & $6.265$ &  $ 9.065 \pm 0.035$ &            $5/5$ &      $0.980$ &     $0.893$ \\
\bottomrule
\end{tabular}
\caption{Clusterability of clusterably-initialized MLPs. See start of appendix~\ref{app:clusterability_data} for details.}
\label{tab:clust_init_mlp_stats}
\end{table}

\begin{table}
\centering
\begin{tabular}{lcccccc}
\toprule
Dataset & Pruning? &  N-cut   &  Dist.\ n-cuts     &  Prop.\ $p<0.02$ &  Train acc.\ &  Test acc.\ \\
\midrule
MNIST   & $\times$ & $10.528$ & $10.658 \pm 0.018$ &         $5/5$    &      $0.998$ &    $0.991$  \\
MNIST   & $\surd$  & $10.148$ & $10.001 \pm 0.039$ &         $0/5$    &      $1.000$ &    $0.993$  \\
Fashion & $\times$ & $10.574$ & $10.662 \pm 0.018$ &         $5/5$    &      $0.982$ &    $0.921$  \\
Fashion & $\surd$  & $10.357$ & $10.034 \pm 0.040$ &         $0/5$    &      $0.995$ &    $0.923$  \\
\bottomrule
\end{tabular}
\caption{Clusterability of clusterably-initialized small CNNs. See start of appendix~\ref{app:clusterability_data} for details.}
\label{tab:clust_init_small_cnn_stats}
\end{table}

\begin{table}
\centering
\begin{tabular}{cccccc}
\toprule
Pruning? &  N-cut   &  Dist.\ n-cuts    &  Prop.\ $p<0.02$ &  Train acc.\ &  Test acc.\ \\
\midrule
$\times$ &  $8.213$ & $3.905 \pm 0.219$ &         $0/5$    &      $0.957$ &     $0.898$ \\
$\surd$  &  $8.212$ & $3.582 \pm 0.229$ &         $0/5$    &      $0.974$ &     $0.902$ \\
\bottomrule
\end{tabular}
\caption{Clusterability of clusterably-initialized VGGs trained on CIFAR-10. See start of appendix~\ref{app:clusterability_data} for details.}
\label{tab:clust_init_vgg_stats}
\end{table}

%% file: appendix_sections/random_datasets.tex
There are two potential dataset-agnostic explanations for why clusterability would increase during training. 
The first is that it increases naturally as a byproduct of applying gradient updates. 
The second is that in order to accurately classify inputs, networks adopt relatively clusterable structure. 
To distinguish between these two explanations, we run three experiments on
a dataset of 28$\times$28 images with i.i.d.\ uniformly random pixel values, associated with random labels between 0 and 9.

In the \textbf{unlearnable random dataset experiment}, we train an MLP on 60,000 random images with default hyperparameters, 10 runs with dropout and 10 runs without.
Since the networks are unable to memorize the labels, this tests the effects of SGD controlling for accuracy.
We compare the n-cuts of the unpruned networks against the distribution of randomly initialized networks to check whether SGD without pruning increases clusterability.
We also compare the n-cuts from the pruned networks against the distribution of n-cuts from shuffles of those networks, to check if SGD increases clusterability in the presence of pruning more than would be predicted purely based on the increase in sparsity.

In the \textbf{kilo-epoch random dataset experiment}, we modify the unlearnable random dataset experiment to remove the pruning and train for 1000 epochs instead of 20,
to check if clusterability simply takes longer to emerge from training when the dataset is random. Note that even in this case, the network is unable to classify the training set better than random.

In the \textbf{memorization experiment}, we modify the random dataset and training method to be more easily learnable. To do this, we reduce the number of training examples to 3,000, train without pruning for 100 epochs and then with pruning for 100 more epochs, and refrain from shuffling the dataset between epochs. As a result, the network is often able to memorize the dataset, letting us observe whether SGD, pruning, and learning can increase clusterability on an arbitrary dataset.

As is shown in table~\ref{tab:ncut-stats-random}, the unlearnable random dataset experiment shows no increase in clusterability before pruning relative to the initial distribution shown in figure~\ref{fig:mlp_init_ncuts}, suggesting that it is not a result of the optimizer alone. 
We see an increase in clusterability after pruning, but no relative clusterability, and absolute clusterability is below that of MLPs trained on MNIST or Fashion-MNIST with no regularization or with dropout displayed in figure~\ref{fig:mlp_clusterability_pruned}.

\begin{table}
\centering
\begin{tabular}{cccccccc}
\toprule
Dropout  & Unp.\ n-cuts       &  N-cuts           & Dist.\ n-cuts     & Prop.\ $p<0.02$ & Train acc.\ \\
\midrule
$\times$ & $10.289 \pm 0.035$ & $9.336 \pm 0.026$ & $9.326 \pm 0.046$ & $0/10$          & $0.101$     \\
$\surd$  & $10.266 \pm 0.041$ & $9.312 \pm 0.040$ & $9.316 \pm 0.042$ & $0/10$          & $0.102$     \\
\bottomrule
\end{tabular}
\caption{Results from the unlearnable random dataset experiment. Reporting as in tables in appendix~\ref{app:clusterability_data}. ``Unp'' is short for unpruned. Accuracies and n-cut distributions are of pruned networks.
}
\label{tab:ncut-stats-random}
\end{table}

The results from the kilo-epoch random dataset experiment are shown in table~\ref{tab:ncut-stats-random-x50}.
The means and standard deviations suggest that even a long period of training caused no increase in clusterability relative to the distribution shown in figure~\ref{fig:mlp_init_ncuts}, 
while pruned networks were not relatively clusterable or as clusterable as those trained on MNIST or Fashion-MNIST plotted in figure~\ref{fig:mlp_clusterability_unpruned}.

\begin{table}
\centering
\begin{tabular}{cccccccc}
\toprule
Dropout  & Unp.\ n-cuts       &  N-cuts           & Dist.\ n-cuts     & Prop.\ $p<0.02$ & Train acc.\ \\
\midrule                                                                                             
$\times$ & $10.267 \pm 0.044$ & $9.332 \pm 0.022$ & $9.331 \pm 0.050$ & $0/10$          & $0.101$     \\
$\surd$  & $10.274 \pm 0.031$ & $9.299 \pm 0.045$ & $9.305 \pm 0.045$ & $0/10$          & $0.101$     \\
\bottomrule
\end{tabular}
\caption{Results from the kilo-epoch random dataset experiment. Reporting as in table~\ref{tab:ncut-stats-random}.
}
\label{tab:ncut-stats-random-x50}
\end{table}

The results of the memorization experiment, shown in table~\ref{tab:ncut-stats-random-memorization}, are different for the networks trained with and without dropout. 
Some networks trained with dropout memorized the dataset, and they appear to be relatively clusterable
Those trained without dropout all memorized the dataset and were all relatively clusterable.
In fact, their degree of clusterability is similar to that of those trained on Fashion-MNIST or MNIST without dropout.
Before the onset of pruning, the n-cuts of the networks trained without dropout were consistently lower than those of randomly initialized networks, as shown in figure~\ref{fig:mlp_init_ncuts}, and the n-cuts of those trained with dropout were at the lower end of the randomly initialized distribution.

\begin{table}
\centering
\begin{tabular}{cccccccc}
\toprule
Dropout  & Unp.\ n-cuts       &  N-cuts           & Dist.\ n-cuts     & Prop.\ $p<0.02$ & Train acc.\ \\
\midrule                                                                                             
$\times$ & $10.093 \pm 0.025$ & $8.746 \pm 0.029$ & $9.154 \pm 0.035$ & $10/10$         & $1.000$     \\
$\surd$  & $10.221 \pm 0.061$ & $8.82 \pm 0.15$   & $8.984 \pm 0.042$ & $6/10$          & $0.213$     \\
\bottomrule
\end{tabular}
\caption{Results from the memorization experiment. Reporting as in table~\ref{tab:ncut-stats-random}.
}
\label{tab:ncut-stats-random-memorization}
\end{table}

Overall, these results suggest that
the training process promotes modularity as a by-product of learning or memorization, and not automatically. Furthermore, we see that dropout fails to promote clusterability when it inhibits memorization.

%% file: appendix_sections/topology-preserving-shuffles.tex
Since SGD alone does not appear to increase clusterability, one might suppose that
the increase in clusterability relative to random networks is due to the pruning producing a clusterable topology, and that the values of the non-zero weights are unimportant. To test this, we compare each trained network to a new distribution: instead of randomly shuffling all elements of each weight matrix, we only shuffle the non-zero elements, thereby preserving the network's topology,

\begin{figure}
\begin{center}
\centerline{\includegraphics[width=0.8\textwidth]{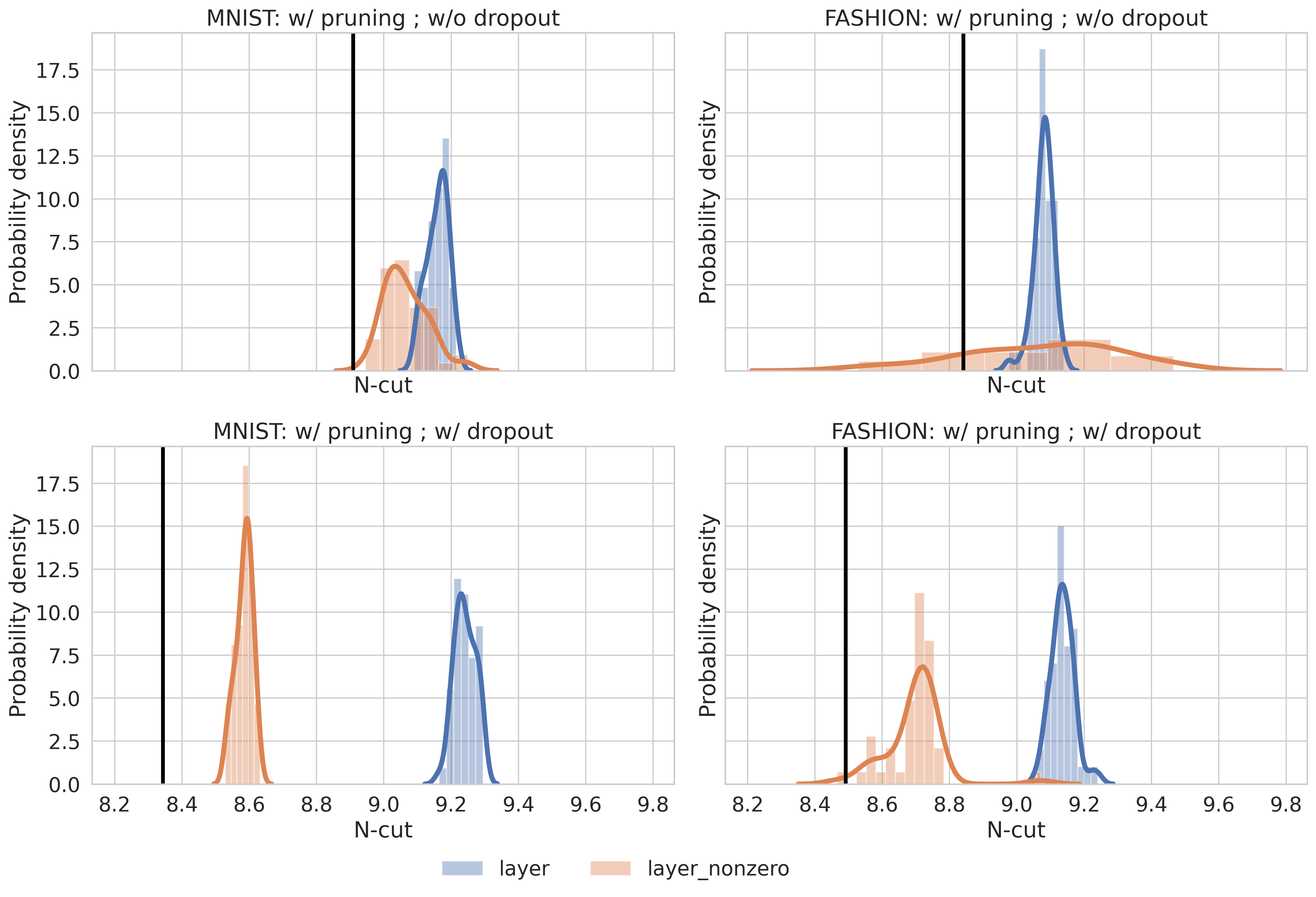}}
\caption{N-cuts of pruned networks trained on MNIST and Fashion-MNIST with and without dropout, compared to the distribution of n-cuts of networks generated by shuffling all elements of each weight matrix (shown in blue, labeled `layer'), as well as the distribution of n-cuts of networks generated by shuffling only the non-zero elements of each weight matrix so as to preserve network topology (shown in orange, labeled `layer-nonzero'). Realized n-cuts are shown as black vertical lines.
Produced by the \texttt{distplot} function of seaborn 0.9.0 \cite{seaborn_0_9_0} with default arguments.}
\label{fig:ncut_dists_layer_nonzero}
\end{center}
\end{figure}

Figure~\ref{fig:ncut_dists_layer_nonzero} shows the n-cuts of some representative networks compared to the distribution of n-cuts of all shuffled networks, and also the distribution of n-cuts of the topology-preserving shuffles. We see three things:
first, that in all cases our networks are more clusterable than would be expected given their topology;
second, that the topology-preserving shuffles are more clusterable than other shuffles,
suggesting that the pruning process is removing the right weights to promote clusterability;
and third, that with dropout, the distribution of topology-preserving shuffles has much lower n-cuts than the distribution of all shuffles.

%% file: appendix_sections/poly_regression.tex
We also train an MLP on a type of polynomial regression task. The inputs to the network are two numbers, $x$ and $y$, each independently normally distributed with mean 0 and variance 1. Each input is associated with a 512-dimensional label. Each label dimension is associated with one of the 512 polynomials in $x$ and $y$ with coefficients in $\{0,1\}$ and exponents in $\{0,1,2\}$. For a given input, each dimension of the label has the value of the corresponding polynomial evaluated at that input.
The network, with 4 hidden layers of 256 neurons each and 512 outputs, is trained to minimize mean square error between its outputs and the label. Test losses range from 0.005 to 0.16 without regularization (with most losses being approximately 0.01), 0.3 when $L_1$ regularization is used, and between 0.07 and 0.3 when $L_2$ regularization is used. Dropout seemed to harm performance and was therefore not used.

Since monomials like $x^2y^2$ can be computed based on monomials $x^2$ and $y^2$, and since each of the 8 valid non-constant monomials appears in 256 of the polynomials, one might expect that the network would adopt a clusterable structure, computing different monomials somewhat independently and combining their results in the output.

\begin{figure}
    \centering
    \includegraphics[width=0.7\columnwidth]{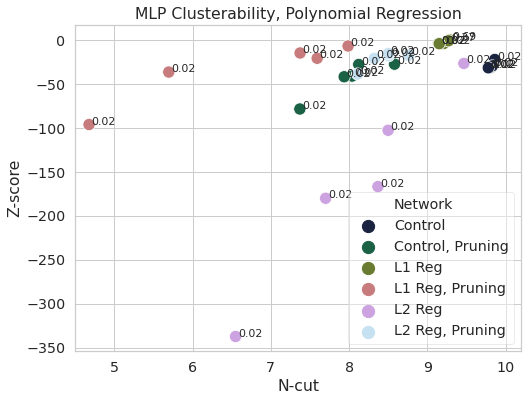}
    \caption{Clusterability of MLPs trained on polynomial regression. Points are labeled with their one-sided $p$-value.}
    \label{fig:mlp_polynomial}
\end{figure}

Results are shown in figure~\ref{fig:mlp_polynomial} and table~\ref{tab:mlp_poly_stats}.
We see that networks trained on this task are consistently relatively clusterable, and that pruning enhances clusterability as usual.

\begin{table}
\centering
\begin{tabular}{lcccccc}
\toprule
Reg.\ method & Pruning? &  N-cut   &  Dist.\ n-cuts     &  Prop.\ $p<0.02$ &  Train loss &  Test loss \\
\midrule
None         & $\times$ &  $9.808$ & $10.187 \pm 0.013$ & $5/5$            &   $ 0.078$  &   $0.046$  \\
None         & $\surd$  &  $8.009$ & $ 9.301 \pm 0.032$ & $5/5$            &   $ 0.025$  &   $0.024$  \\
$L_1$        & $\times$ &  $9.210$ & $ 9.278 \pm 0.031$ & $3/5$            &   $ 0.350$  &   $0.319$  \\
$L_1$        & $\surd$  &  $6.665$ & $ 8.468 \pm 0.059$ & $5/5$            &   $ 0.211$  &   $0.192$  \\
$L_2$        & $\times$ &  $8.115$ & $ 9.867 \pm 0.012$ & $5/5$            &   $ 0.175$  &   $0.140$  \\
$L_2$        & $\surd$  &  $8.438$ & $ 9.800 \pm 0.067$ & $5/5$            &   $ 0.101$  &   $0.078$  \\
\bottomrule
\end{tabular}
\caption{Clusterability of MLPs trained on polynomial regression. See start of appendix~\ref{app:clusterability_data} for details. Losses averaged over 5 runs.}
\label{tab:mlp_poly_stats}
\end{table}

%% file: appendix_sections/mixture_results.tex
We initially hypothesized that modularity is a result of different regions of the network processing different types of input. 
To test this,
we developed mixture datasets composed of two original datasets.
These datasets are either of the `separate' type, where one original dataset has only classes 0 through 4 included and the other has classes 5 through 9 included; or the `overlapping' type, where both original datasets contribute examples of all classes. The datasets that we mix are MNIST and LINES, which consists of 28$\times$28 images of white vertical lines on a black background, labeled with the number of vertical lines.

If modularity were a result of different regions of the network specializing in processing different types of information, we would expect that networks trained on mixture datasets would have n-cuts lower than those trained on either constituent dataset.
This is not what we observe.

Table~\ref{tab:mixture-no-dropout} shows n-cuts and accuracies for networks trained with pruning but without dropout, while table~\ref{tab:mixture-dropout} shows the same for networks trained with pruning and dropout. Networks trained on mixtures between LINES and MNIST have n-cuts intermediate between those trained on LINES and those trained on MNIST.
That being said, the artificial nature of the LINES dataset, as well as the low test accuracy of networks trained with dropout on LINES (seemingly implying that dropout in this case increased the degree of overfitting), put the generalizability of these results into question.

\begin{table}
\centering
\begin{tabular}{lcccc}
\toprule
Dataset            & N-cuts            & Mean train acc.\ & Mean test acc.\ \\
\midrule
MNIST              & $8.880 \pm 0.049$ & $1.000$          & $0.984$         \\
LINES              & $7.361 \pm 0.096$ & $1.000$          & $1.000$         \\
LINES-MNIST        & $7.955 \pm 0.096$ & $1.000$          & $0.991$         \\
LINES-MNIST-SEP    & $8.26 \pm 0.19$   & $1.000$          & $0.994$         \\
\bottomrule
\end{tabular}
\caption{N-cuts and accuracies for networks trained without dropout. LINES-MNIST refers to the dataset where each class has data from both LINES and MNIST, while LINES-MNIST-SEP refers to the dataset where classes 0-4 have examples from LINES and classes 5-9 have examples from MNIST. Each row presents statistics for 5 networks. The ``N-cuts'' column shows the mean and standard deviation over 5 networks.}
\label{tab:mixture-no-dropout}
\end{table}

\begin{table}
\centering
\begin{tabular}{lcccc}
\toprule
Dataset            & N-cuts            & Mean train acc.\ & Mean test acc.\ \\
\midrule
MNIST              & $8.350 \pm 0.046$ & $0.967$          & $0.979$         \\
LINES              & $6.85 \pm 0.16$   & $0.912$          & $0.296$         \\
LINES-MNIST        & $6.933 \pm 0.094$ & $0.873$          & $0.637$         \\
LINES-MNIST-SEP    & $7.779 \pm 0.072$ & $0.984$          & $0.894$         \\
\bottomrule
\end{tabular}
\caption{N-cuts and accuracies for networks trained with dropout. Notation as in table~\ref{tab:mixture-no-dropout}.}
\label{tab:mixture-dropout}
\end{table}

%% file: appendix_sections/clust_grad.tex
In this section, we show how to compute the gradient of the $k$\ts{th} eigenvalue, $\lambda_k$, of $\Lnorm$ with respect to each weight matrix $W^s$. First, we remember that we can equivalently consider that eigenvalue as the $k$\ts{th} eigenvalue of $\Lsym$, and that the gradient of $\lambda_k$ with respect to $\Lsym$ is the outer product of the corresponding eigenvector $u_k$ with itself, $u_k u_k^\top$. Therefore, we only need compute the gradient of $\Lsym$ with respect to each weight matrix $W^s$. We will then have
\begin{align*}
    \frac{\partial \lambda_k}{\partial W^s_{n,m}} &= \sum_{ij} (u_k)_i (u_k)_j \frac{\partial (\Lsym)_{i,j}}{\partial W^s_{n,m}}.
\end{align*}

Since $\Lsym = I - D^{-1/2}AD^{-1/2}$, $(\Lsym)_{i,j} = \delta_{i,j} - d_i^{-1/2} d_j^{-1/2} A_{i,j}$, where $\delta_{i,j}$ is the function that is 1 if $i=j$ and 0 otherwise. Furthermore, for symmetry, we write $d_i = (1/2) (\sum_k A_{i,k} + \sum_l A_{l,i})$. This gives the partial derivatives
\begin{align*}
    \frac{\partial d_i}{\partial A_{i,i}} &= 1 \\
    \frac{\partial d_i}{\partial A_{i,j}} &= \frac{1}{2} \\
    \frac{\partial d_i}{\partial A_{k,i}} &= \frac{1}{2}, 
\end{align*}
where $j,k \neq i$.

From here on out, we need not consider derivatives with respect to $(\Lsym)_{i,i}$ or $A_{i,i}$, since $(\Lsym)_{i,i} = A_{i,i} = 0$ always by definition, because nodes don't have edges to themselves. We can now compute derivatives of $\Lsym$ with respect to $A$:
\begin{align*}
    \frac{\partial (\Lsym)_{i,j}}{\partial A_{i,j}} &= - d_i^{-1/2} d_j^{-1/2} + \frac{1}{4} A_{i,j} \left( d_i^{-3/2}d_j^{-1/2} + d_i^{-1/2}d_j^{-3/2} \right) \\
    \frac{\partial (\Lsym)_{i,j}}{\partial A_{i,l}} &= \frac{1}{4} A_{i,j} d_i^{-3/2} d_j^{-1/2}  \\
    \frac{\partial (\Lsym)_{i,j}}{\partial A_{k,j}} &= \frac{1}{4} A_{i,j} d_i^{-1/2} d_j^{-3/2}  \\
    \frac{\partial (\Lsym)_{i,j}}{\partial A_{k,l}} &= 0,
\end{align*}
where $i \neq k$ and $j \neq l$.

Next, let $e(n,s)$ be the index of neuron $n$ of layer $s$, and let the weight matrices be $W^1$ through $W^S$. It's then the case that 
\begin{equation*}
    \frac{\partial A_{e(n,s),e(m,s+1)}}{\partial W^s_{n,m}} = \frac{\partial A_{e(m, s+1),e(n,s)}}{\partial W^s_{n,m}} = \textrm{sgn}(W^s_{n,m}),
\end{equation*}
and $\partial A_{i,j} / \partial W^s_{n,m} = 0$ for all other $i$ and $j$, by the definition of how $A$ is constructed from the weight matrices, where $\textrm{sgn}(x)$ is the sign of $x$: $1$ if $x > 0$, $-1$ if $x < 0$, and $0$ if $x = 0$.

Therefore, combining all the equations above, we have:
\begin{align*}
    \frac{\partial (\Lsym)_{i,j}}{\partial W^s_{n,m}} &= \textrm{sgn}(W^s_{n,m}) \left( \frac{\partial (\Lsym)_{i,j}}{\partial A_{e(n,s),e(m,s+1)}} + \frac{\partial (\Lsym)_{i,j}}{\partial A_{e(m,s+1),e(n,s)}} \right) \\
    &= \sgn(W^s_{n,m}) \left( \delta_{i,e(n,s)} \times \frac{1}{4} A_{i,j} d_i^{-3/2} d_j^{-1/2} + \delta_{j,e(m,s+1)} \times \frac{1}{4}A_{i,j} d_i^{-1/2}d_j^{-3/2} - \delta_{i,e(n,s)} \delta_{j,e(m,s+1)} d_i^{-1/2} d_j^{-1/2} \right. \\
    &\phantom{= \sgn(W^s_{n,m})} \left. {} + \delta_{i,e(m,s+1)} \times \frac{1}{4} A_{i,j} d_i^{-3/2} d_j^{-1/2} + \delta_{j,e(n,s)} \times \frac{1}{4}A_{i,j} d_i^{-1/2}d_j^{-3/2} - \delta_{i,e(m,s+1)} \delta_{j,e(n,s)} d_i^{-1/2} d_j^{-1/2} \right).
\end{align*}

\noindent{This concludes our derivation.}

%% file: appendix_sections/clust_reg_norm.tex
In a network with ReLU activation functions, it is possible to scale the inputs of a hidden neuron by a positive constant $c$ while scaling the outputs by $1/c$ without changing the function the network computes. Since having more weights close to zero makes it easier for clustering algorithms to cut low-weight edges while preserving high-weight edges, regularizing for clusterability will induce these transformations.
However, they do not make the network any more modular in any real sense, which is our aim.

Therefore, when applying this regularizer, we manually apply this transformation after each gradient step to enforce the invariant that the norm of the input weights to each hidden neuron should be $\sqrt{2}$, chosen to maintain the benefits of He initialization \cite{he2015delving, kumar2017weight}.
This is detailed in algorithm~\ref{alg:clust_reg_norm}.

\begin{algorithm}[tb]
\caption{Normalization for MLP Clusterability Regularization}
\label{alg:clust_reg_norm}
\begin{algorithmic}
\STATE {\bfseries input:} Weight matrices $W_1, \dotsc, W_m$, bias vectors $b_1, \dotsc, b_m$
\FOR{hidden layers $i$ from 1 to $m-1$}
\FORALL{neurons $n$ in the layer}
\STATE form $v$ by concatenating the $n$\ts{th} column of $W_i$ (that is, the weights that feed into neuron $n$) and the $n$\ts{th} entry of $b_i$
\STATE set $x := \sqrt{\sum_j v_j^2}$
\IF{$x \neq 0$}
\STATE multiply $n$\ts{th} column of $W_i$ by $\sqrt{2}/x$
\STATE multiply $n$\ts{th} entry of $b_i$ by $\sqrt{2}/x$
\ENDIF
\STATE multiply $n$\ts{th} row of $W_{i+1}$ by $x/\sqrt{2}$
\ENDFOR
\ENDFOR
\STATE {\bfseries return:} New weight matrices $W_1, \dotsc, W_m$ and bias vectors $b_1, \dotsc, b_m$
\end{algorithmic}
\end{algorithm}